\documentclass[conference]{IEEEtran}
\IEEEoverridecommandlockouts
\usepackage{amsmath,amssymb,amsfonts}
\usepackage{graphicx}
\usepackage{textcomp}
\usepackage{xcolor}
\usepackage{adjustbox}
\usepackage{cite}
\def\BibTeX{{\rm B\kern-.05em{\sc i\kern-.025em b}\kern-.08em
   T\kern-.1667em\lower.7ex\hbox{E}\kern-.125emX}}

\usepackage[font=small,labelfont=bf]{caption}

\usepackage{times}
\usepackage{epsfig}
\usepackage{graphicx}
\usepackage{amsmath}
\usepackage{amssymb}

\usepackage{tabularx}
\usepackage[font={small}]{caption}
\usepackage[font={small}]{subcaption}
\usepackage{float}
\usepackage{multirow}
\usepackage{makecell}

\usepackage[numbers]{natbib}

\usepackage{nicefrac} 
\usepackage{upgreek} 
\usepackage{isomath} 

\usepackage{units} 
\usepackage{algpseudocode}
\usepackage{mathtools}
\usepackage{footnote}
\usepackage{url}
\usepackage{adjustbox}
 
\usepackage[acronym,shortcuts,acronymlists={hidden}]{glossaries}

\usepackage{hyperref}
\usepackage{cleveref}

\newacronym{SVM}{SVM}{support vector machine}
\newacronym{CNN}{CNN}{convolutional neural networks}
\newacronym{BoF}{BoF}{Bag of Features}
\newacronym{ORN}{ORN}{Oriented Response Network}
\newacronym{ORConv}{ORConv}{Oriented Response Convolution}
\newacronym{A-ORConv}{A-ORConv}{Averaged Oriented Response Convolution}
\newacronym{ARF}{ARF}{Active Rotating Filter}
\newacronym{A-ARF}{A-ARF}{Averaged Active Rotating Filter}
\newacronym{IORN}{IORN}{Improved Oriented Response Network}

\newcommand{\mat}[1]{\mathbf{#1}} 
\newcommand{\hmat}[1]{\mathbf{\hat{#1}}} %
\newcommand{\tmat}[1]{\mathbf{\tilde{#1}}} %
\DeclareMathAlphabet\mathbfcal{OMS}{cmsy}{b}{n}

\newcommand{\footref}[1]{%
    $^{\ref{#1}}$%
}

\makeatletter

\AtBeginDocument{
    \setlength{\belowdisplayskip}{4pt} \setlength{\belowdisplayshortskip}{2pt}
    \setlength{\abovedisplayskip}{4pt} \setlength{\abovedisplayshortskip}{2pt}
}
\makeatletter
\let\origsection\section
\renewcommand\section{\@ifstar{\starsection}{\nostarsection}}

\newcommand\nostarsection[1]
{\sectionprelude\origsection{#1}\sectionpostlude}

\newcommand\starsection[1]
{\sectionprelude\origsection*{#1}\sectionpostlude}

\newcommand\sectionprelude{%
  \vspace{-1pt}
}

\newcommand\sectionpostlude{%
  \vspace{-2pt}
}

\let\origsubsection\subsection
\renewcommand\subsection{\@ifstar{\starsubsection}{\nostarsubsection}}

\newcommand\nostarsubsection[1]
{\subsectionprelude\origsubsection{#1}\subsectionpostlude}

\newcommand\starsubsection[1]
{\subsectionprelude\origsubsection*{#1}\subsectionpostlude}

\newcommand\subsectionprelude{%
  \vspace{-3pt}
}

\newcommand\subsectionpostlude{%
  \vspace{-3pt}
}

\let\origsubsubsection\subsubsection
\renewcommand\subsubsection{\@ifstar{\starsubsubsection}{\nostarsubsubsection}}

\newcommand\nostarsubsubsection[1]
{\subsubsectionprelude\origsubsubsection{#1}\subsubsectionpostlude}

\newcommand\starsubsubsection[1]
{\subsubsectionprelude\origsubsubsection*{#1}\subsubsectionpostlude}

\newcommand\subsubsectionprelude{%
  \vspace{-1pt}
}

\newcommand\subsubsectionpostlude{%
  \vspace{-1pt}
}

\usepackage{xspace}

\makeatletter
\DeclareRobustCommand\onedot{\futurelet\@let@token\@onedot}
\def\@onedot{\ifx\@let@token.\else.\null\fi\xspace}

\def\eg{\emph{e.g}\onedot} 
\def\ie{\emph{i.e}\onedot}

\makeatother

\begin{document}

\title{Fashion Landmark Detection and Category Classification for Robotics}

\author{{ Thomas Ziegler$^{1,2}$, Judith Butepage$^{2}$, Michael C. Welle$^{2}$, Anastasiia Varava$^{2}$, Tonci Novkovic$^{1}$ and Danica Kragic$^{2}$ } \thanks{$^{1}$Thomas Ziegler and Tonci Novkovic,  ETH Eidgenössische Technische Hochschule Zürich {\tt\small zieglert@ethz.ch, tonci.novkovic@mavt.ethz.ch}}  \thanks{$^{2}$ Thomas Ziegler, Judith Butepage, Michael Welle, Anastasiia Varava, and Danica Kragic, KTH Royal Institute of technology         {\tt\small butepage,mwelle,varava,dani@kth.se}}  }
        


\maketitle

\begin{abstract}
    
    Research on automated, image based identification of clothing categories and fashion landmarks has recently gained significant interest due to its potential impact on areas such as robotic clothing manipulation, automated clothes sorting and recycling, and online shopping. Several public and annotated fashion datasets have been created to facilitate research advances in this direction. In this work, we make the first step towards leveraging the data and techniques developed for fashion image analysis in vision-based robotic clothing manipulation tasks.
    We focus on techniques that can generalize from large-scale fashion datasets to less structured, small datasets collected in a robotic lab. Specifically, we propose training data augmentation methods such as elastic warping, and model adjustments such as rotation invariant convolutions to make the model generalize better. Our experiments demonstrate that our approach outperforms state-of-the art models with respect to clothing category classification and fashion landmark detection when tested on previously unseen datasets. Furthermore, we present experimental results on a new dataset composed of images where a robot holds different garments, collected in our lab. 
\end{abstract}


\section{Introduction}
\label{sec:introduction}

As the interest for fashion items increases in online shopping and e-commerce, the need for automated image analysis in the fashion industry is growing. This application area requires many tasks to be automatized, such as clothing category classification, fashion landmark detection, image retrieval and similarity based recommendations. Following the creation of large-scale fashion datasets \cite{Liang2015DeepHuman, Liu2016DeepFashion, Liu2016Landmarks}, significant progress has been made in fashion image analysis. Deep learning based models have achieved significant performance gain in clothing category classification \cite{Huang2015CrossDomain, Liu2016DeepFashion, Lu2017Fully, Corbiere2017Leveraging,Ge2019Deepfashion2}, item recommendation \cite{Liu2016DeepFashion, Ma2017Towards, Han2017LearningFashion}, and retrieval \cite{Kiapour2015Where, Liu2016DeepFashion}.

However, for robotic clothing manipulation,  the collection of large-scale datasets proves to be more difficult. Robotic clothing manipulation includes tasks such as clothing category classification \cite{Willimon2013AnewApproach, Ramsia2013FINDD, Li2014Recognition, Li2014RealTime, Stria2018Classification} and tasks that require fashion landmark detection, such as grasp point detection \cite{Ramsia2016A3Ddescriptor, Corona2018Active}, folding \cite{Doumanoglou2016Folding, garcia2020benchmarking}, sorting \cite{Sun2017Singleshot}, unfolding \cite{Doumanoglou2014Autonomous,Doumanoglou2014Active}, and dressing \cite{garcia2020benchmarking}. Compared to retail applications, the vast majority of existing work on clothing category classification in robotics uses custom datasets for evaluation. These datasets are often limited in the number of images and contain only a small number of different categories.

\begin{figure}
    \centering
    \includegraphics[width=0.325\linewidth]{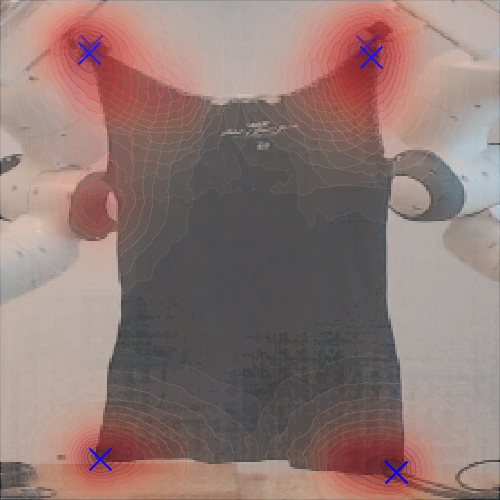}
    \includegraphics[width=0.325\linewidth]{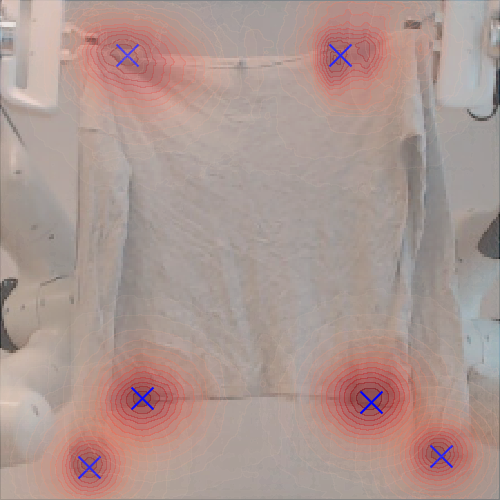}
    \includegraphics[width=0.325\linewidth]{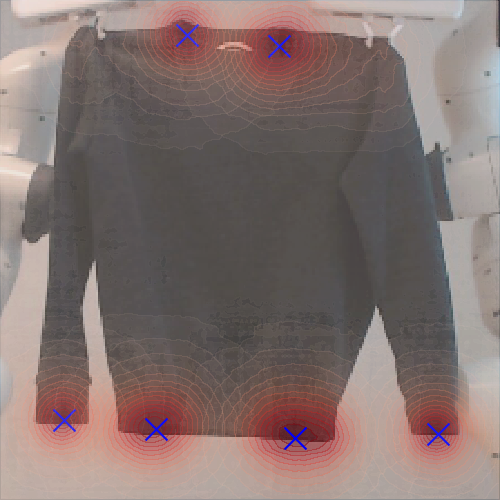}
    \vspace{0.1cm}
    \centering
    \includegraphics[width=0.325\linewidth]{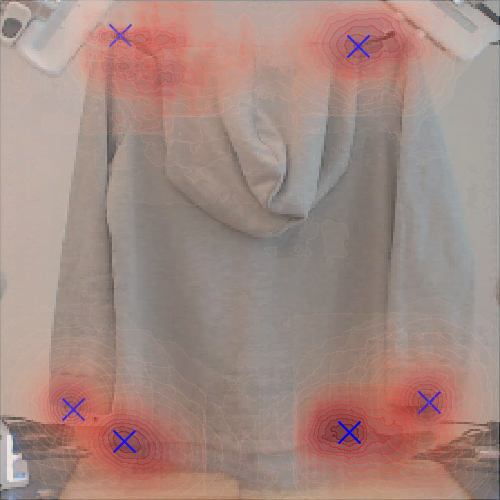}
    \includegraphics[width=0.325\linewidth]{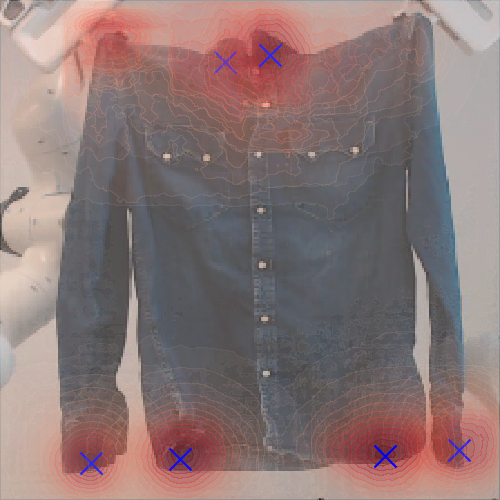}
    \includegraphics[width=0.325\linewidth]{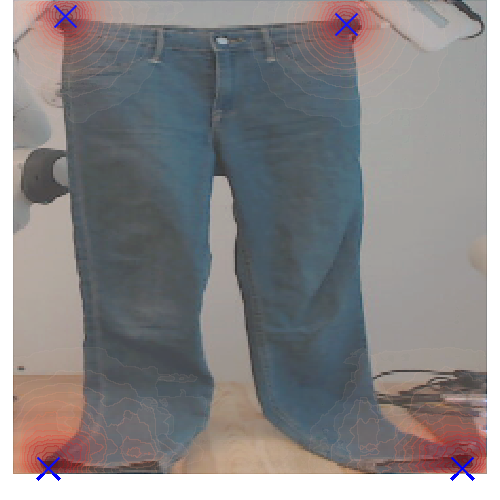}
  \caption{Example images of the landmark localization on the six categories of our in-lab dataset. The categories are from top left to bottom right: Tank, Tee, Sweater, Hoody, Jacket, Jeans. Robot arms are visible in the images. The predicted heatmaps are shown in red and the blue crosses denote the selected maximum values.}
  \label{fig:inlab_dataset}
  \vspace{-0.5cm}
\end{figure}

In this work, we identify two tasks that are common to both retail and robotic applications, namely clothing category classification and fashion landmark detection. While the fashion industry often considers structured data, such as a human wearing clothes facing the camera, the data in robotic applications is less structured and can contain images of upside-down, crumpled clothing items. We build upon the progress made in fashion image analysis and propose a network architecture and training procedure on a large-scale fashion dataset DeepFashion \cite{Liu2016DeepFashion}. Our model is capable to generalize well to the noisy, poorly controlled conditions encountered in robotic clothing manipulation. 
DeepFashion dataset contains over 280000 images of clothing separated into 46 categories and annotated with $4\sim8$ landmarks per item. In order to resemble the more challenging clothing configurations encountered in robotic manipulation, we introduce \textit{elastic warping}, a novel image augmentation method. It uses random displacement fields to create authentic looking clothing configurations. Our proposed model incorporates rotation invariance and attention mechanisms in order to handle difficult configurations faced in robotic manipulation, such as random orientation.  

The performance of our model is evaluated extensively on different publicly available datasets. One of these is a small-scale dataset created by us, which contains real world images typically encountered in a robotic manipulation task, \ie where robot arms are visible in the image as shown in Figure \ref{fig:inlab_dataset}. In contrast to other, state-of-the-art methods, our approach is able to generalize to new, much smaller datasets without additional fine-tuning. 
We illustrate a potential application scenario of our model by performing landmark detection on a garment that is being folded by a robot. We demonstrate that the landmarks are successfully detected even when the garment is  partially occluded by the robot (see supplementary video\footref{fn:code}). 

The landmark detection is very stable despite the robot occluding parts of the garment during the manipulation.
Therefore we believe the proposed method is a first step for robotic clothing manipulation tasks that require basic visual information such as category classification and landmark detection.

The contributions of our work is fourfold:

\emph{(i)} We propose a novel deep learning based network for clothing classification and landmark prediction specific for robot manipulation. To the best of our knowledge, this is the first deep learning based computer vision model for clothing manipulation trained solely on RGB images. \emph{(ii)} We introduce \textit{elastic warping} for landmark detection, a new data augmentation method that is capable of resembling more challenging clothing configurations which are not encountered in standard datasets. \emph{(iii)} We provide extensive experimental results on different datasets which identify a lack of generalization to novel datasets of other state-of-the-art fashion networks. \emph{(iv)} We created a small dataset, containing real world images of clothing in a realistic robotic manipulation environment. Additionally, we annotated landmark position in the CTU dataset \cite{wagner2013ctu}.  The annotations, the in-lab dataset and the implementation are publicly available \footnote{ \label{fn:code} \url{https://github.com/ThomasZiegler/Fashion_Landmark_Detection_and_Category_Classification}}.

\section{Related Work}
\label{sec:related_work}

We survey the work from the computer vision and robotic communities related to our work.

\subsection{Fashion Networks}
\label{ssec:fashion_networks}
Fashion image analysis has drawn increased attention in the field of computer vision due to its impact on e-commerce and online shopping. Several deep learning networks evolved from this trend for the different analysis tasks of clothing recognition \cite{Huang2015CrossDomain, Liu2016DeepFashion, Lu2017Fully, Corbiere2017Leveraging,Ge2019Deepfashion2},  recommendation \cite{Liu2016DeepFashion, Ma2017Towards, Han2017LearningFashion},  retrieval \cite{Kiapour2015Where}, and fashion landmark localization \cite{Liu2016DeepFashion, Liu2016Landmarks, Yan2017Unconstrained, Wang2018Attentive, Liu2018DFAnalysis, Ge2019Deepfashion2}. In our work, the focus is on category classification and landmark localization.

 \citeauthor{Liu2016DeepFashion} \cite{Liu2016DeepFashion} propose a multi-branch network for simultaneous classification, retrieval and landmark localization. In \cite{Liu2016Landmarks}, the same authors propose a combination of three cascaded networks for a gradual refinement of landmark localization. \citeauthor{Yan2017Unconstrained} \cite{Yan2017Unconstrained} use recurrent spatial transformers in combination with selected dilated convolutions to predict landmarks in unconstrained scenes. More recently, \citeauthor{Wang2018Attentive} \cite{Wang2018Attentive} proposed a deep fashion grammar network for combined clothing category classification and landmark localization. The network encodes two attention mechanisms: landmark-aware and category-driven attention. A similar network was proposed by \citeauthor{Liu2018DFAnalysis} \cite{Liu2018DFAnalysis} which has an increased resolution in the predicted heatmaps for the landmarks and uses a unified attention mechanism instead of two separate streams.

\subsection{Computer vision for robotic clothing manipulation}
\label{ssec:cloth_detection_networks}
Robotic clothing manipulation is a well established research area with pioneer work going back more than twenty years \cite{Hamajima1996Planning}. It can be seen as a pipeline of several consecutive steps to bring an item of clothing from an unknown state into a desired one (\eg folded or sorted) \cite{Doumanoglou2016Folding}. A broad overview over methods used for visual grasp point localization, classification and state recognition is given in \cite{Jimenez2017Visual}. 

Compared to the structured data used for retail applications, the image data used in the robotics community is of a different nature. The items are either lying in a spread or crumpled state on a flat surface \cite{Willimon2013AnewApproach, Ramsia2013FINDD, Sun2016Recognising, Sun2017Singleshot} or they are in a hanging state when grasped by a robotic gripper \cite{Kita2004ADeformable,Li2014Recognition, Mariolis2015Pose, Kampouris2016MultiSensorial, Gabas2016RobotAided, Corona2018Active, Stria2018Classification}. 

A major difference to vision applications in the fashion industry is that the robotics community has mostly focused on task specific, handcrafted feature extraction, such as edges and corners \cite{willimon2011model} and wrinkles \cite{alenya2012characterization}. Additionally, due to the 3D nature of the manipulation task, the use of physics and volumental simulators is more common in robotics \cite{Kita2009Clothes, Li2014RealTime}. 

Some recent methods \cite{Mariolis2015Pose, Kampouris2016MultiSensorial, Gabas2016RobotAided, Corona2018Active} use \ac{CNN} instead of handcrafted features to classify hanging items of clothing. All models are shallow, containing a few convolutional layers followed by a few fully connected layers. \citeauthor{Stria2018Classification} \cite{Stria2018Classification} use a \ac{CNN} to create a global feature vector from depth maps. The \ac{CNN} is trained on a large dataset with common 3D objects. For classification on a smaller clothing dataset, the CNN is used to extract features which are passed to a Support Vector Machine.

Our proposed network has a similar architecture as the networks proposed in \cite{Wang2018Attentive, Liu2018DFAnalysis}, but it has been extended to a more challenging clothing configurations encountered in robotic applications. To the best of our knowledge, this is the first deep learning based network designed as part of a robotic clothing manipulation pipeline that only uses RGB images.

\section{Method}
\label{sec:method}
In this section we formulate the problem and introduce two augmentation methods to resemble clothing configurations encountered in robotic clothing classification and landmark localization tasks. Finally we give a detailed description of the proposed network.  

\subsection{Problem Formulation}
\label{ssec:problem_formulation}
Given an image $\mat{I}$ the goal is to simultaneously predict the landmark locations $\mat{L}$ and category classification $\mat C$. The landmarks are defined as $\mat L= \left\{(x_k, y_k)\right\}_{k=1}^{n_L}$, where $(x_k, y_k)$ is the $k$th pixel coordinate position in $\mat{I}$ and $n_L$ the total number of landmarks per image. 
 
The category classification $\mat C \in [0,1]^{n_C}$ satisfies  $\sum_{i=1}^{n_C} C_i = 1$, where $n_C$ is the number of categories depending on the used dataset.  

\subsection{Image augmentation}
\label{ssec:preprocessing}

In order to make the available fashion datasets more representative for robotic applications we propose two types of data augmentation: image rotation and \textit{elastic warping}. One challenge is to augment an image together with its landmarks. We define the image before transformation as input image $\mat{I}$ and the image after the transformation as transformed image $\tmat I$. In both cases $w, h$ stand for the width and height of the image respectively. 

 The transformation can be represented as a mapping of the pixels, $\forall (\tilde{x}, \tilde{y}) \in [1,w]\times [1,h]$:
 \begin{align}
  \label{eq:tranformation}
  \begin{gathered}
    \tmat I(\tilde{x},\tilde{y})= \mat I\big(x(\tilde{x},\tilde{y}), y(\tilde{x},\tilde{y})\big).
  \end{gathered}
\end{align}
Where $x,y$ are the pixel location in the input image $\mat{I}$ and $\tilde{x},\tilde{y}$ the pixel location in the transformed image $\tmat I$. The clothing landmark locations $\mat L= \left\{(x_k, y_k)\right\}_{k=1}^{n_L}$ are a set of $n_L$ specific pixel coordinates in the input image $\mat{I}$. 

When $x(\tilde{x},\tilde{y})$ and/or $y(\tilde{x},\tilde{y})$ are non-integer, interpolation is needed. We apply the commonly used bilinear interpolation \cite{Simard2003Best} in such a case.

\subsubsection{Rotation}
\label{sssec:rotation}
A rather simple but powerful augmentation is image rotation. Rotating images with a small angle is often used to increase the performance in classification and/or detection tasks \cite{Shorten2019}. When items of clothing lie on the ground or on a table, they can be in any orientation. We hence randomly sample an angle $\theta$ in the range $[0,2\pi]$ for each rotation.

 
 \subsubsection{Elastic Warping}
\label{sssec:elastic_warping}
To resemble the distortion of loose items of clothing, we propose an elastic warping method. The method is similar to the elastic deformation proposed in \cite{Simard2003Best} but is further extended to produce realistic, task-specific images and to allow for landmark detection. 

The deformation is created by generating two random displacement fields $\mat \Delta \mat x(\tilde{x}, \tilde{y})$ and $\mat \Delta \mat y(\tilde{x}, \tilde{y})$. The whole augmentation is performed in four steps:

\textit{First:} Sample $n_S$ pixel positions uniformly in the transformed image:  $\mathbfcal{S} = \{(\tilde{x}_i, \tilde{y}_i) \}_{i=1}^{n_S}$.

\textit{Second:} For each pixel location in $\forall (\tilde{x}_i, \tilde{y}_i) \in \mathbfcal{S}$ sample a random displacement from a uniform distribution $\mathcal{U}(-\alpha, \alpha)$:
\begin{equation}
  \begin{split}
    \mat \Delta \mat x(\tilde{x}_i, \tilde{y}_i) \sim \mathcal{U}(-\alpha, \alpha), \; \mat \Delta \mat y(\tilde{x}_i, \tilde{y}_i) \sim \mathcal{U}(-\alpha, \alpha).
  \end{split}
\end{equation}
All other entries in the displacement fields are set to $0$.

\textit{Third:} Convolve the two displacement fields with a Gaussian filter $\mat G, \ \forall (\tilde{x}, \tilde{y}) \in [1,w] \times [1,h]$ :
\begin{align}
  \begin{split}
    \mat \Delta \bar{\mat x}(\tilde{x}, \tilde{y}) =  \mat \Delta \mat x(\tilde{x}, \tilde{y}) * \mat G(\tilde{x}, \tilde{y}) 
  \end{split} \\
  \begin{split}
   \mat  \Delta \bar{\mat y}(\tilde{x}, \tilde{y}) =  \mat \Delta \mat y(\tilde{x}, \tilde{y}) * \mat G(\tilde{x}, \tilde{y})  
 \end{split}
\end{align}
where $*$ denotes the convolution operator.   and $\mat G(\tilde{x}, \tilde{y})$ is a Gaussian filter with variance parameter $\sigma$.

\textit{Fourth:} Use the smoothed displacement field to create the transformed image, $ \forall (\tilde{x}, \tilde{y}) \in [1,w] \times [1,h]$ :
\begin{equation}
  \begin{gathered}
    \tmat I(\tilde{x},\tilde{y}) = \mat I\big(\underbrace{\tilde{x} + \mat \Delta  \bar{\mat x}(\tilde{x},\tilde{y})}_{x(\tilde{x},\tilde{y})}, \underbrace{\tilde{y} + \mat \Delta \bar{\mat y}(\tilde{x},\tilde{y}}_{y(\tilde{x},\tilde{y})})\big)   
  \end{gathered}
\end{equation}

\begin{figure}
    \centering
    \includegraphics[width=0.325\linewidth]{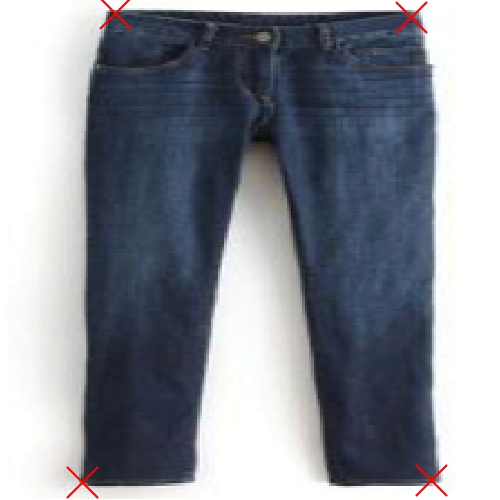}
    \includegraphics[width=0.325\linewidth]{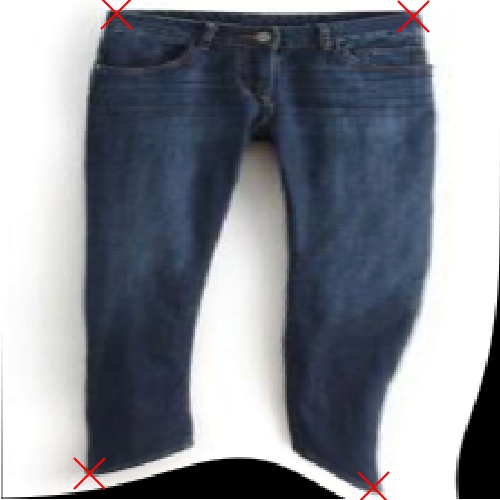}
    \includegraphics[width=0.325\linewidth]{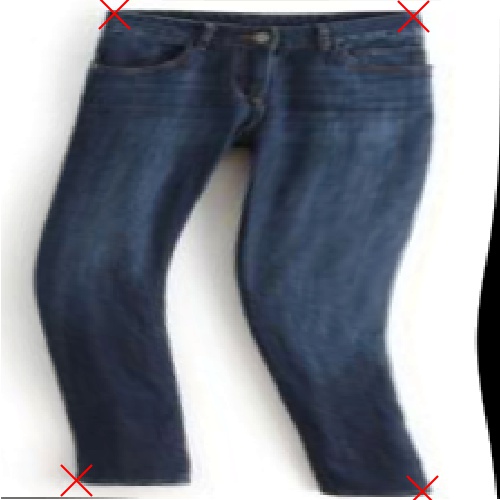}
    \vspace{0.1cm}
    \includegraphics[width=0.325\linewidth]{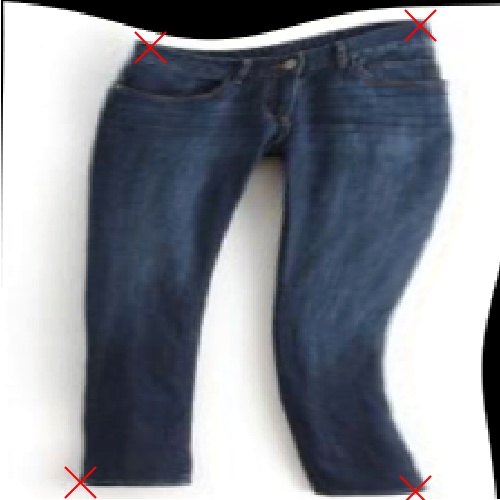}
    \includegraphics[width=0.325\linewidth]{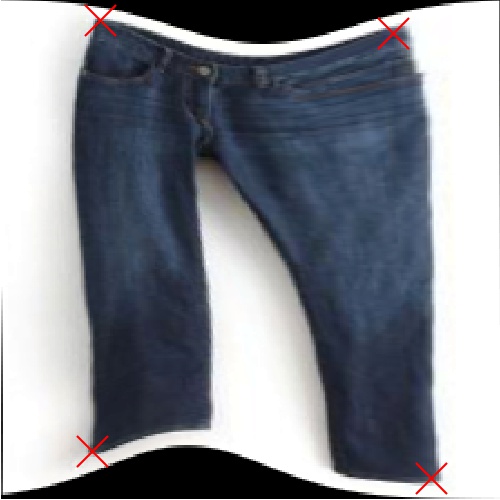}
    \includegraphics[width=0.325\linewidth]{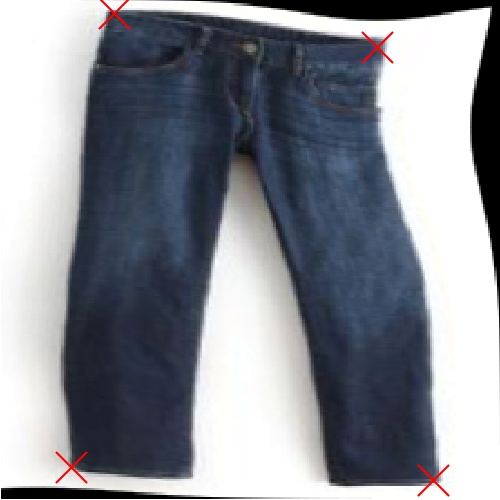}
    \caption{Example images of our proposed elastic warping with $n_S=3$, $\alpha=500$ and $\sigma=40$. Top left is the original image, all others are transformed versions using different random seeds. Each landmark is marked with a red cross.}
  \label{fig:elastic_warping}
  \vspace{-0.5cm}
\end{figure}

The strength of the distortion can be adjusted by the number of initially displaced pixels $n_S$, the scaling of the uniform distribution $\alpha$ and the smoothness of the Gaussian filter $\sigma$. We use $n_S=3$, $\alpha=500$ and $\sigma=40$ in our experiments. Figure \ref{fig:elastic_warping} shows some examples when using this configuration.

\noindent \textbf{Landmark warping}
The displacement fields indicate where a pixel in the transformed image was located in the input image. Due to the random nature of these fields no inverse exists. That means that it is not trivial to know if/where the pixels of the input image are found in the transformed image. This is problematic for landmark warping, since their location is only defined for the input image. In the following we describe an efficient method for retrieving the landmark position in the transformed image.

For every landmark position $\mat L_k = (x_k, y_k)$ we find $n$ possible pixels in the transformed image $\tmat I$ which originated at or near the position of the landmark in the input image $\mat I$: 
\begin{align}
  \mathbfcal{X} = \operatorname*{argmin-\mathnormal{n}}\limits_{\forall (\tilde{x}, \tilde{y}) \in [1,w] \times [1,h]}  \operatorname{sort} |\tilde{x} + \mat \Delta  \bar{\mat x}(\tilde{x},\tilde{y}) -x_k | \\
  \mathbfcal{Y} = \operatorname*{argmin-\mathnormal{n}}\limits_{\forall (\tilde{x}, \tilde{y}) \in [1,w] \times [1,h]}  \operatorname{sort} |\tilde{y} + \mat \Delta  \bar{\mat y}(\tilde{x},\tilde{y}) -y_k |,
\end{align}
where $\operatorname*{argmin-\mathnormal{n}}$ returns the $n$ smallest values from a sorted set. Note that both $\mathbfcal{X}$ and $\mathbfcal{Y}$ contain coordinate pairs $(\tilde{x},\tilde{y})$. The value of $n$ depends on the image size and the chosen parameters $n_S, \alpha,$ and $\sigma$ in the elastic warping. We use $n=200$ in our experiments.
To get the transformed landmark $\tmat L_k$  we need to find the coordinate pair $(\tilde{x}^*,\tilde{y}^*)$ that is either present in both $\mathbfcal{X}$ and $\mathbfcal{Y}$ or the coordinate pair in $\mathbfcal{X}$ with closest neighbor in $\mathbfcal{Y}$.

We use the fact that the pixel coordinates are unique integer values and create a hash table for all coordinate pairs in one set. In the following, one can search for each pair in the other set if a key exist in the hash table, which reduces time complexity for existing coordinated pairs to $\mathcal{O}(n)$.

If the hash table does not return a valid value, no exact match exists in $\mathbfcal{X}$ and $\mathbfcal{Y}$. In this case, one can create a kd-tree for all coordinate pairs in $\mathbfcal{Y}$ and use kd-tree search \cite{Moore1990Efficient} to find the nearest neighbor for the coordinate pairs in $\mathbfcal{X}$. The average search time for kd-tree search is $\mathcal{O}(n \log n)$.


\subsection{Network Architecture}
\label{sec:network_architecture}
The main network architecture is loosely based on the VGG-16 \cite{Simonyan2015VeryDeep} network structure similar to the networks proposed in \cite{Liu2018DFAnalysis,Wang2018Attentive}. The structure can be seen in Figure \ref{fig:networka}. Compared to the base VGG-16 network, several structural changes are included: rotation invariance layers (Section \ref{ssec:rotationinv}), a landmark localization branch (Section \ref{ssec:landmark_localization_branch}) and attention branches for classification (Section \ref{ssec:attention_branch}). As many components are inspired by prior work, we focus the discussion on the main components and direct interested readers to the Appendix for detailed network structure descriptions. 


\subsubsection{Rotation invariance}
\label{ssec:rotationinv}
Orientation variation occurs more often in robotic clothing classification images than they do in fashion images. 
In order to account for this, we replace the $2$D convolution in the \textit{conv\textsl{1}} to \textit{conv\textsl{4}} layers with \acp{A-ORConv}. They produce enriched feature maps with the orientation information explicitly encoded \cite{Wang2018IORN}.

\begin{figure}
\begin{subfigure}[b]{1\linewidth}
  \includegraphics[trim=80 0 0 0, width=0.9\linewidth]{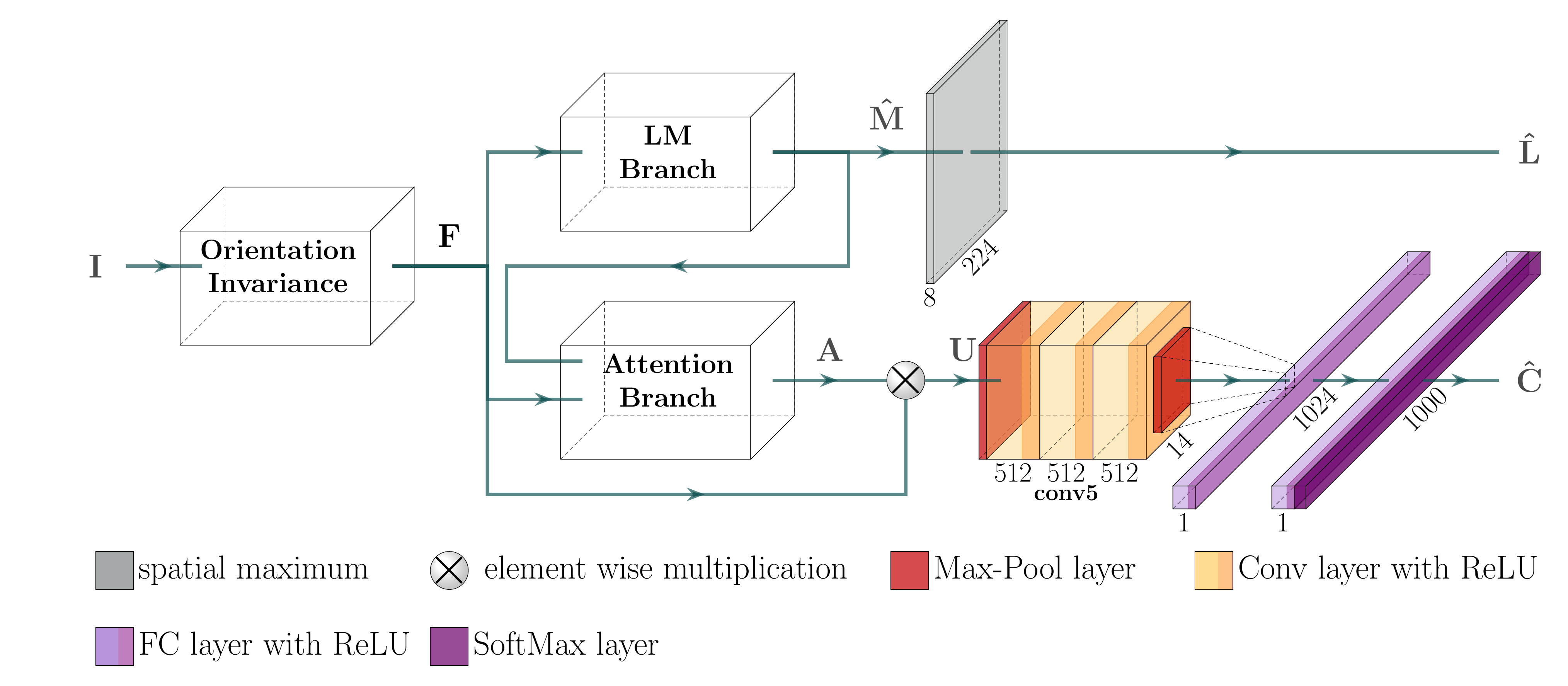}
  \caption{Overall network structure} \label{fig:networka}
  \par\medskip 
  \includegraphics[width=0.9\linewidth]{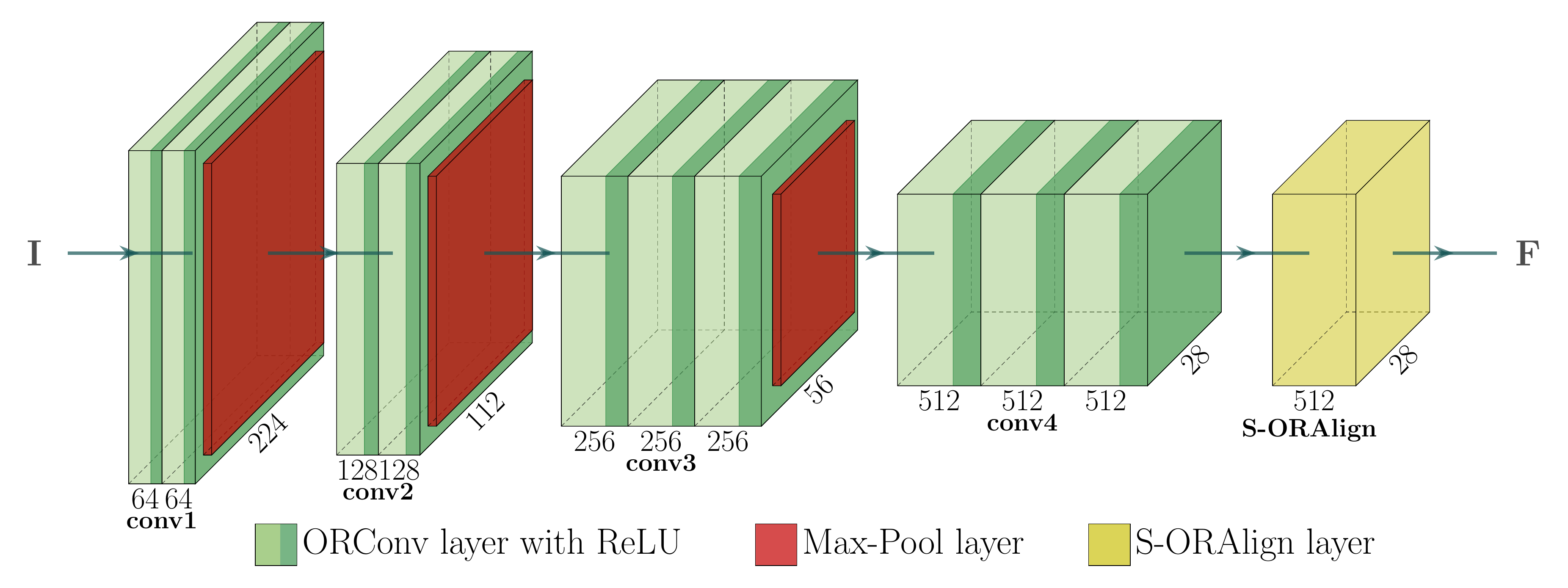}
  \caption{Rotation invariance encoder} \label{fig:networkb}
  \par\medskip 
 \includegraphics[width=0.9\linewidth]{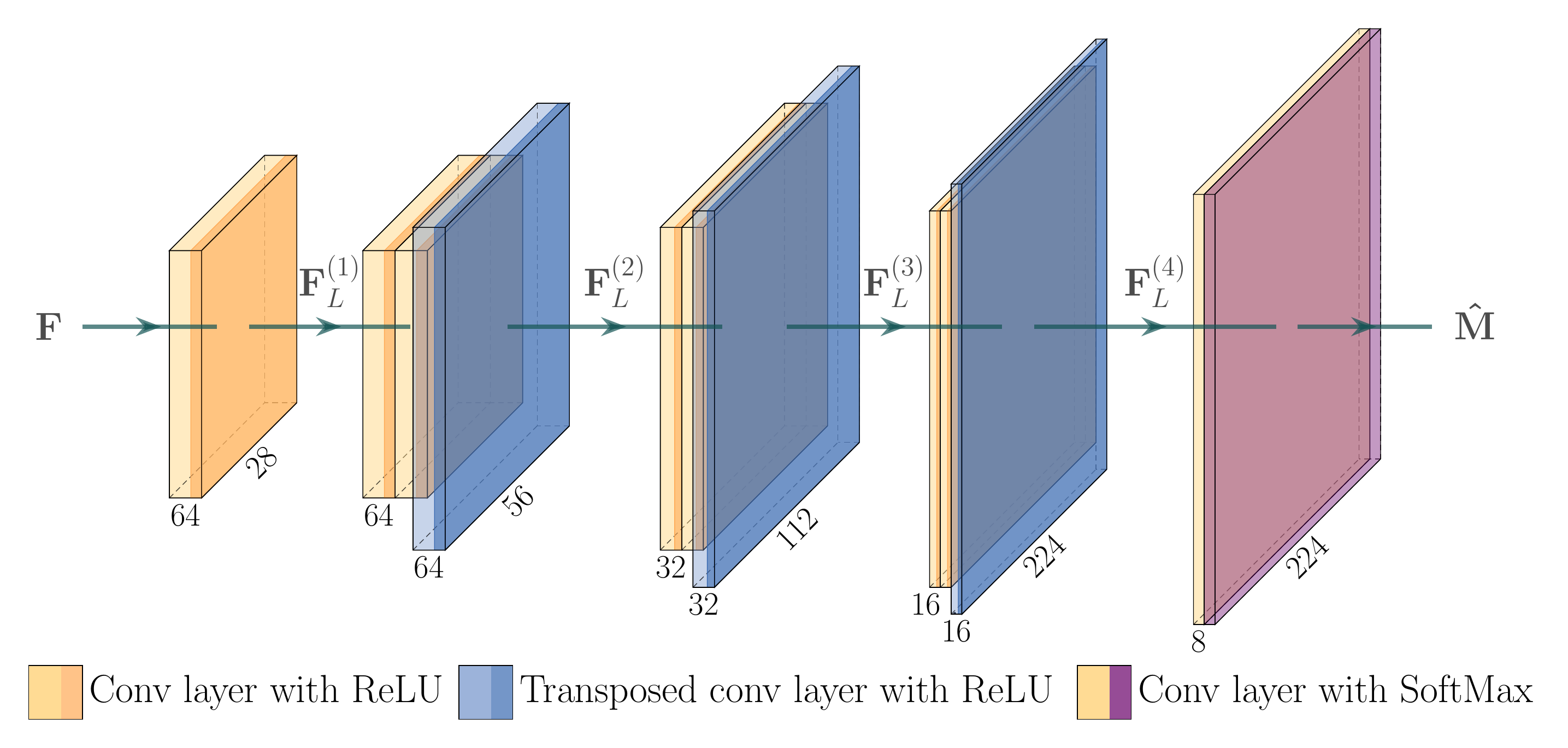}  
  \caption{Landmark (LM) localization branch} \label{fig:networkc}
  \par\medskip 
  \includegraphics[trim=100 0 0 0, width=0.9\linewidth]{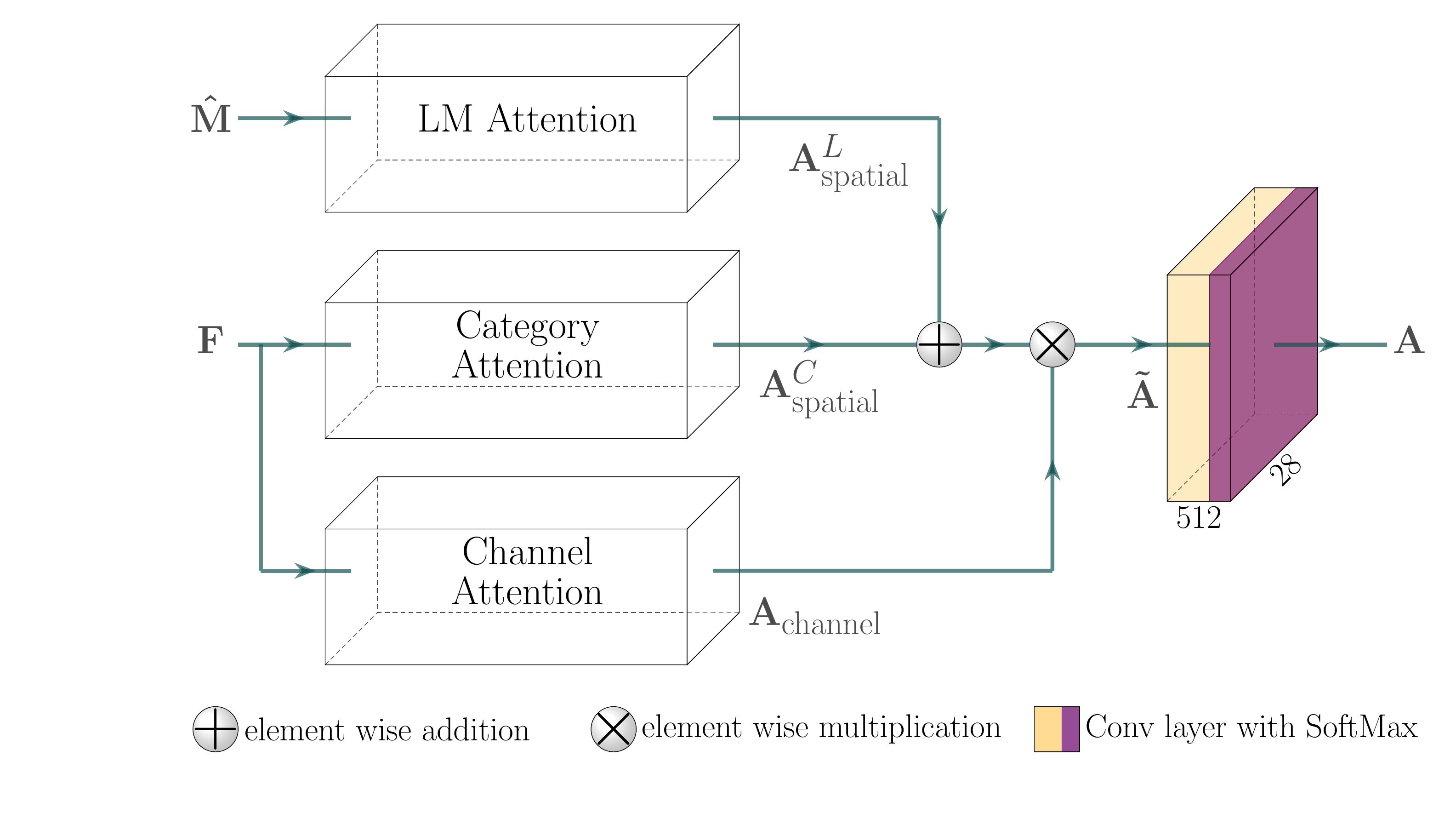}
  \caption{Attention branch} \label{fig:networkd}
  \par\medskip 
   \includegraphics[width=0.9\linewidth]{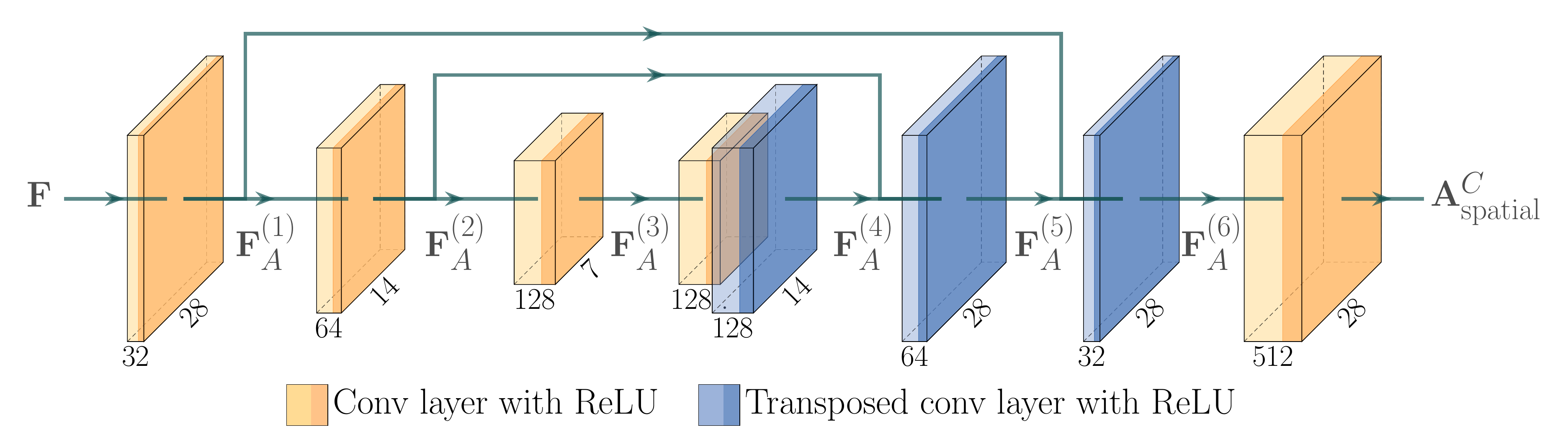}
  \caption{Category aware spatial attention} \label{fig:networke}
  \par\medskip 
  \includegraphics[width=0.9\linewidth]{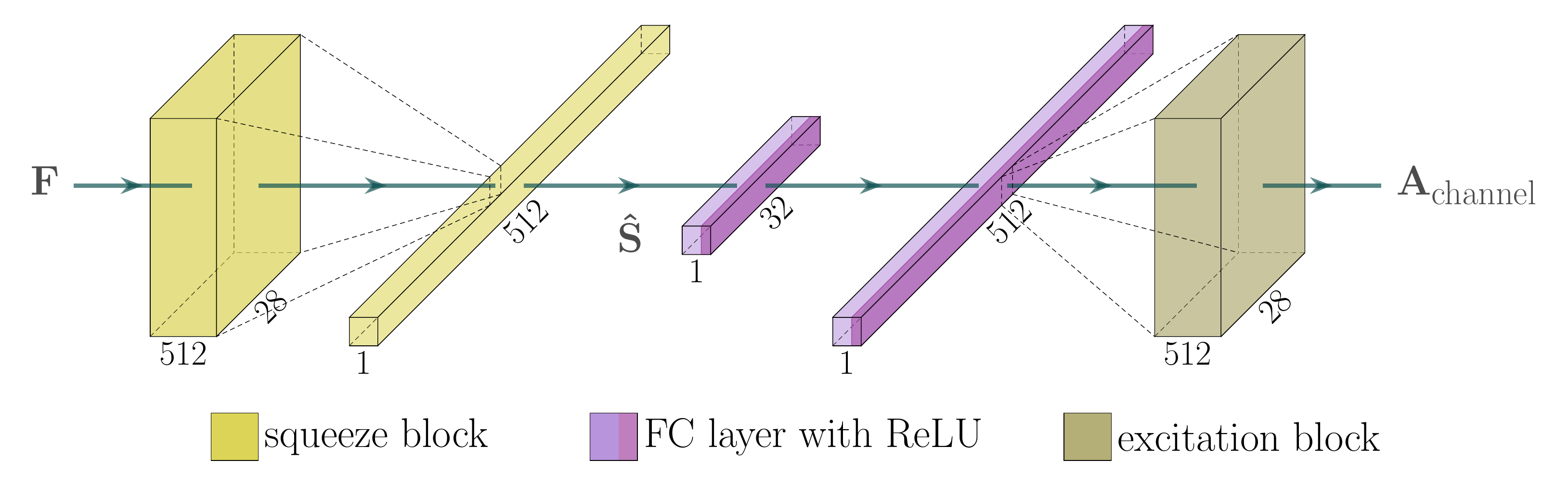}
  \caption{Channel attention} \label{fig:networkf}
\end{subfigure}

\caption{The different components of our model.}\label{fig:network}

\end{figure}

In our network (Figure \ref{fig:networkb}), we use the \acp{A-ORConv} with four orientation channels (\ie $N\!=\!4$). We use the same filter size and the same number of total channels when replacing the standard $2$D convolution in the \textit{conv\textsl{1}} to \textit{conv\textsl{4}} layers. This means that the effective number of parameters of the \acp{A-ORConv} is only a quarter of the normal convolution blocks. 
In order to create rotation invariant features, a Squeeze-ORAlign (S-ORAlign) layer \cite{Wang2018IORN} is used to find the main response channel. 

\subsubsection{Landmark Localization Branch}
\label{ssec:landmark_localization_branch}
The landmark localization branch is the same as proposed in \cite{Liu2018DFAnalysis}. The branch structure is depicted in Figure \ref{fig:networkc}.

The landmark localization branch can be trained separately from the classification. Given that the extracted feature map \textbf{F} is of dimension $w_f\times h_f\times n_L$, where $w_f$ and $h_f$ are width and height of the feature map and $n_L$ is the number of landmarks, the ground-truth heatmap and the predicted heatmap for the $k$th landmark can be denoted by $\mat M_k \in [0,1]^{w_f\times h_f}$ and $\hmat M_k \in [0,1]^{ w_f \times h_f}$ respectively. The landmark localization branch is trained using pixel-wise mean square differences,
\begin{align}
  \mathcal{L}_{\text{LM}} =  \sum_{i=1}^{n_B} \sum_{k=1}^{n_L} \sum_{x=1}^{w_f} \sum_{y=1}^{h_f} \| \mat M_k^i(x,y) - \hmat M_k^i(x,y) \|_2^2,
\end{align}
where $n_B$ is the total number of training samples. The ground-truth heatmap $\mat M_k^i$ is generated by adding a $2$D Gaussian filter at the corresponding location $\mat L^i_k$. Given a sample $i$ the predicted coordinates for the $k$th landmark $\hmat L^i_k$ corresponds to the maximal value in the predicted heatmap,

\begin{equation}
  \hmat L^i_k \in \underset{(x,y) \in [1,w_f] \times [1,h_f]}{\mathop{\text{argmax}}} \hmat M^i_k(x,y).
\end{equation}
If there is more than one maximum per landmark one of them is chosen at random.

\subsubsection{Attention Branch}
\label{ssec:attention_branch}
The attention branch can be seen as a union of \textit{spatial} attention \cite{Jaderberg2015Spatial} and \textit{channel} attention \cite{Hu2018Squeeze}. The attention learns a saliency weight map  $\mat A$.
Inspired by the proposed attention modules in \cite{Wang2018Attentive} the spatial attention itself contains two types of attention, a landmark attention $\mat A^L_\text{spatial}$ and a category attention $\mat A^C_\text{spatial}$. 
Thus, the attention branch is designed as a three branch unit; two branches for the spatial attention $\mat A_\text{spatial}^L, \mat A_\text{spatial}^C$ (Figure \ref{fig:networkd}) and one for the channel attention $\mat A_\text{channel}$ (Figure \ref{fig:networkf}). These are combined in a factorized manner as explained below,
\begin{equation}
  \label{eq:attention_factorization}
  \mat A = (\mat A^L_\text{spatial}+\mat A^C_\text{spatial}) \times \mat A_\text{channel}.
\end{equation}
\vspace{4pt}

\noindent \textbf{Spatial Attention - Landmark}
Clothing landmarks represent functional regions of clothing and provide useful information about an item. The predicted heatmaps $\{\hmat M_k\}_{k=1}^{n_L}$ are used to guide attention to the functional clothing regions. The weight map is created by downsampling the predicted heatmaps which is followed by a max-pooling operation. This attention is learned in a supervised manner since it is directly derived from the predicted heatmaps. 

\noindent \textbf{Spatial Attention - Category}
Since the landmark attention only covers corner points of a clothing item, an additional spatial attention is used that focuses more on the clothing center. The category attention (Figure \ref{fig:networke}) is modeled using an U-Net structure \cite{Ronneberger2015UNet}. 
The model learns by itself which regions of an image are important. This is in contrast to our landmark attention, where the groundtruth heatmaps $\mat M$, which resemble the landmark attention, are provided during training.  

\noindent \textbf{Channel Attention}
The channel attention (Figure \ref{fig:networkf}) is implemented via a Squeeze-and-Excitation block \cite{Hu2018Squeeze}. First a \textit{squeeze} operation creates  $\mat S$, an embedding of the global distribution of the channel-wise feature responses. Then an \textit{excitation} operation is performed on the channel wise aggregated feature map to create the channel attention. Following the proposal in \cite{Hu2018Squeeze} a bottleneck is created using two fully-connected layers, with a reduction rate $r$.

\noindent \textbf{Factorization}
The factorization (Figure \ref{fig:networkd}) is performed by multiplying the channel-wise feature responses in the spatial attention with the corresponding channel weights,
To refine the attention, an additional $1\!\times\!1$ convolution layer is added afterwards. This is motivated by the fact that the spatial and channel attention are not mutually exclusive but with co-occurring complementary relationship \cite{Li2018Harmonious}.
\subsubsection{Output architecture}

Given $\mat A$, we weight the S-ORAlign features $\mat F$,  $\mat U = (\mat 1+\mat A) \circ \mat F$, where $\circ$ denotes the Hadamard product and $\mat 1$ is a tensor. Hence, features where $\mat A(\cdot, \cdot, \cdot) \in [-1, 0)$ are reduced and features where $\mat A(\cdot, \cdot, \cdot) \in (0, 1]$ are increased. Our attention incorporates semantic information and global information into the network helping to focus on important regions in the images. The features $\mat U$ are then fed in to the \textit{conv\textsl{5-1}} layer. The rest of the network follows the VGG-16 structure. 

%
%

\section{Experiments}
\label{sec:experiments}
This section describes several different experiments to evaluate the performance of the proposed network and learning procedure. The section starts with a description of the different datasets we use, followed by the descriptions of the individual experiments and results.

\subsection{Datasets}
\label{ssec:datasets}

In the following, we introduce all datasets used for training and/or evaluation.

\subsubsection{DeepFashion dataset}
\label{sssec:deepfashion_dataset}
The DeepFashion: Category and Attribute Prediction Benchmark (DeepFashion dataset\footnote{\url{http://mmlab.ie.cuhk.edu.hk/projects/DeepFashion/AttributePrediction.html}}) \cite{Liu2016DeepFashion} is a large collection of fashion images. It offers 289222 images collected from the Google image search engine and from shopping websites. The dataset contains 8 different landmarks (\ie \textit{left/right collar}, \textit{left/right sleeve}, \textit{left/right waistline}, and \textit{left/right hem}), 46 clothing categories and 1000 clothing attributes. For each image a bounding box is provided. We use this dataset to train our network with our proposed augmentation methods and then perform inference on small-scale datasets.

\subsubsection{CTU Color and Depth Image Dataset of Spread Garments}
\label{sssec:ctu_dataset}
The CTU Color and Depth Image Dataset of Spread Garments (CTU dataset\footnote{\url{https://github.com/CloPeMa/garment_dataset)}}) \cite{wagner2013ctu} is designed for testing and benchmarking garment segmentation and recognition. The dataset contains 1372 images of size $1280\times1024$ taken from a bird's eye perspective. There are 17 different items divided into 9 categories. Compared to the DeepFashion dataset the clothing items can be in any orientation and they contain not only flat spread but also wrinkled items. We manually labeled the landmark positions in each image. We use this dataset to train our network and evaluate its performance on more challenging clothing configurations typical in robotics. We also use it to evaluate the effect of our proposed augmentation methods when purely trained on the DeepFashion dataset.

\subsubsection{In-Lab Dataset}
\label{sssec:robot_dataset}
As a first step towards generalizing classification and landmark detection results to images that are typical for robotic tasks, we created a small dataset. It contains 117 images from 6 different clothing categories (\ie Tank, Tee, Sweater, Hoody, Jacket, Jeans). Each item is hold by two robotic arms at predefined grasping points (\ie shoulders and waist). This state can be reached with an autonomous unfolding process as proposed in \cite{Doumanoglou2014Autonomous, Doumanoglou2014Active}. The images are of size $960\times720$. Each item of clothing is captured in 9 different configurations of the robotic arms, such that they can overlap with the bounding box around the item. Furthermore, the background is not uniform and is partially cluttered. We annotated the images with the same landmarks as in the DeepFashion dataset and extracted a similar bounding box around each item. We use this dataset to evaluate the performance of our network on previously unseen items in a realistic lab environment.

\begin{table*}[!ht]
\footnotesize

      \begin{adjustbox}{width=1\textwidth}
		\begin{tabular}{l|cccccccc|c}
      Methods (Trained on DF) & L.Collar & R.Collar & L.Sleeve & R.Sleeve & L.Waistline & R.Waistline & L.Hem & R.Hem & Avg. \\
      \hline 
      \citeauthor{Liu2018DFAnalysis} \cite{Liu2018DFAnalysis}  & 0.5056 & \textbf{0.4810} & 0.3288 & 0.2623 & 0.4908 & \textbf{0.4665} & 0.4047 & 0.4774 & 0.4272 \\
      
      Ours & \textbf{0.4972} & 0.4835 & \textbf{0.2846} & \textbf{0.2055} & \textbf{0.4870} & 0.4677 & 0.4069 & 0.4727 & \textbf{0.4131} \\ 
       
      \citeauthor{Liu2018DFAnalysis} \cite{Liu2018DFAnalysis} EW  & 0.5096 &	0.4995 &	0.3314 &	0.2626 &	0.4992 &	0.4730 &	\textbf{0.4063} &	\textbf{0.4698} &	0.4314 \\
      
      Ours EW & 0.5194 &	0.5204 &	0.3538 &	0.2601 &	0.4935 &	0.5251 &	0.4185 &	0.4805 &	0.4464 \\ 
      
     \hline
     \citeauthor{Liu2018DFAnalysis} \cite{Liu2018DFAnalysis} R &  0.0947 & 0.1004 & 0.0814 & 0.0670 & 0.1215 & \textbf{0.1018} & 0.2196 & 0.2177 & 0.1255\\
      
      Ours R & 0.1056 & 0.1075 & 0.0763 & 0.0708 & 0.1133 & 0.1206 & 0.1756 & 0.1526 & 0.1153 \\
      
      \citeauthor{Liu2018DFAnalysis} \cite{Liu2018DFAnalysis} R \& EW & \textbf{0.0863} &	\textbf{0.0880} &	0.0775 &	0.0717 &	\textbf{0.1030} &	0.1265 &	0.2039 &	0.1860 &	0.1179 \\
      
      Ours R \& EW & 0.0999 &	0.0949 &	\textbf{0.0639} &	\textbf{0.0581} &	0.1039 &	0.1151 &	\textbf{0.1557} &	\textbf{0.1474} &	\textbf{0.1047} \\  
       
      \hline

      Methods (Trained on CTU)  & L.Collar & R.Collar & L.Sleeve & R.Sleeve & L.Waistline & R.Waistline & L.Hem & R.Hem & Avg. \\
      \hline 
      \citeauthor{Liu2018DFAnalysis} \cite{Liu2018DFAnalysis} & 0.0560 & 0.0484 & 0.0473 & 0.0572 & 0.0473 & 0.0560 & 0.1010 & 0.0929 & 0.0632 \\
      
      Ours & 0.0500 & 0.0801 & 0.0790 & 0.0745 & 0.0590 & 0.0713 & 0.0749 & 0.0853 & 0.0719 \\
      
      \citeauthor{Liu2018DFAnalysis} \cite{Liu2018DFAnalysis} EW  &0.0447 &	0.0442 &	0.0447 &	0.0481 &	0.0612 &	0.0826 &	0.0860 &	0.0780 &	0.0612 \\
       
      Ours EW & \textbf{0.0260} &	\textbf{0.0267} &	\textbf{0.0319} &	\textbf{0.0262} &	\textbf{0.0311} &	\textbf{0.0359} &	\textbf{0.0620} &	\textbf{0.0548} &	\textbf{0.0368} \\ 
      
     \hline
     
      \citeauthor{Liu2018DFAnalysis} \cite{Liu2018DFAnalysis} R  & 0.0299 & 0.0314 & 0.0289 & 0.0335 & 0.0560 & 0.0402 & 0.0539 & 0.0460 & 0.0400\\
      
      Ours R & \textbf{0.0181} & \textbf{0.0194} & \textbf{0.0253} & \textbf{0.0192} & 0.0374 & 0.0382 & \textbf{0.0314} & 0.0383 & 0.0284 \\ 
      
     \citeauthor{Liu2018DFAnalysis} \cite{Liu2018DFAnalysis} R \& EW  & 0.0295 &	0.0277 &	0.0370 &	0.0403 &	0.0350 &	0.0561 &	0.0483 &	0.0509 &	0.0406 \\
      
      Ours R \& EW & 0.0199 &	0.0248 &	0.0348 &	0.0244 &	\textbf{0.0274} &	\textbf{0.0204} &	0.0334 &	\textbf{0.0276} &	\textbf{0.0266} \\  
       
      \hline

    \end{tabular}
    \end{adjustbox}
  \caption{Results on CTU dataset for landmark localization with different augmentation methods, when trained on the DeepFashion (DF) dataset (top) and in the CTU dataset (bottom). The values represent the normalized error (NE). Best results are marked in bold}
	\label{table:CTU_LM_DF} \vspace{-0.2cm}
\end{table*}

\setlength{\tabcolsep}{3.5pt}
\begin{table*}[ht!]
  \begin{adjustbox}{width=1\textwidth}
  \begin{tabular}{l|cccccccccccccccc|cc}
      Methods (Trained on DF) & \multicolumn{2}{c}{Bluse} & \multicolumn{2}{c}{Hoody} & \multicolumn{2}{c}{Pants} & \multicolumn{2}{c}{Polo} & \multicolumn{2}{c}{Polo-Long} & \multicolumn{2}{c}{Skirt} & \multicolumn{2}{c}{Tshirt} & \multicolumn{2}{c|}{Tshirt-Long} &\multicolumn{2}{c}{Overall}\\
              & top-1 & top-3 & top-1 & top-3 & top-1 & top-3 & top-1 & top-3 & top-1 & top-3 & top-1 & top-3 & top-1 & top-3  & top-1 & top-3 & top-1 & top-3\\
      \hline
      \citeauthor{Liu2018DFAnalysis} \cite{Liu2018DFAnalysis}  & 35.00 & 50.00 & 31.58 & 52.63  & 33.33 & 52.28 & 33.33 & 52.28 & 33.33 & 52.28 & 36.84 & 52.63 & 33.33 & 52.28 & 33.33 & 52.28 & 33.74 & 52.15 \\
      Ours  &  20.00 & 50.00 & 21.05 & 47.37 & 19.05 & 47.62 & 19.05 & 47.62 & 19.05 & 47.62 & 21.05 & 52.63 & 19.05 & 47.62 & 19.05 & 47.62 & 19.63 & 48.47\\
      \citeauthor{Liu2018DFAnalysis} \cite{Liu2018DFAnalysis}  R  & 45.00 & 70.00 & 42.11 & 68.42 & 42.86 & 71.43 & 42.86 & 71.43 & 42.86 & 71.28 & 42.11 & 68.42 & 42.86 & 71.43 & 42.86 & 71.43 & 42.94 & 70.77  \\
       Ours R &  \textbf{85.00} & \textbf{90.00} & \textbf{84.21} & \textbf{89.47} & \textbf{85.71} & \textbf{90.48} & \textbf{85.71} & \textbf{90.48} & \textbf{85.71} & \textbf{90.48} & \textbf{84.21} & \textbf{89.47} & \textbf{85.71} & \textbf{90.48} & \textbf{85.71} & \textbf{90.48} & \textbf{85.28} &\textbf{90.18}  \\
      
       \citeauthor{Liu2018DFAnalysis} \cite{Liu2018DFAnalysis}  R \& EW   & 55.00 & 75.00 & 52.63 & 73.68 & 57.14 & 76.19 & 57.14 & 76.19 & 57.14 & 76.19 & 52.63 & 73.68 & 57.14 & 76.19 & 57.14 & 76.19 & 55.83 & 76.46 \\
       Ours  R \& EW & 80.00 & \textbf{90.00} & 73.68 & \textbf{89.47} & 76.19 & \textbf{90.48} & 76.19 & \textbf{90.48} & 76.19 & \textbf{90.48} & 78.95 & \textbf{89.47} & 76.19 & \textbf{90.48} & 76.19 & \textbf{90.48} & 76.69 &\textbf{90.18} \\
      \hline 
      
    \end{tabular}
    \end{adjustbox}
    \caption{Results on CTU dataset category classification with different augmentation methods, when trained on the DeepFashion (DF) dataset. Best results marked in bold.}
	\label{table:CTU_cat_DF}\vspace{-0.2cm}
\end{table*} 
\setlength{\tabcolsep}{6pt}


\subsection{Pretraining on the DeepFashion Dataset}
\label{ssec:experiment_deepfashion}

In this section we describe the pre-training details for the DeepFashion Dataset. Experimental results on this dataset can be found in the Appendix.

\subsubsection{Experimental Setup}
We use the same settings as \cite{Liu2016DeepFashion, Wang2018Attentive, Liu2018DFAnalysis} for training and evaluation. In total 209222 images are used for training and 40000 images for validation. The final evaluation is performed on the remaining 40000 images. We use the normalized error (NE) \cite{Liu2016Landmarks} as the landmark localization error measure. This is the $l_2$ distance between the predicted and groundtruth landmark in normalized coordinates. For the category and attribute classification top-$k$ classification accuracy is used.

The images are cropped using the provided bounding boxes. We train our model with and without our proposed data augmentation steps whereas the evaluation is always performed without augmentation. For implementation details, please see the Appendix.
 
\subsection{Experiments on CTU Dataset}
\label{ssec:experiment_CTU}

In this section we present experiments on the  CTU Dataset. We perform two different types of experiments on the CTU dataset.
In the first experiment, we analyze the inference performance of our network, solely trained on the entire DeepFashion dataset. Special interest lies in the proposed data augmentation methods, since the clothing configurations differ from the DeepFashion images. In the second experiment we evaluate the performance of our network when trained and evaluated on the CTU dataset.

\subsubsection{Experimental Setup}

It is important to note that the DeepFashion dataset has more than five times the number of categories than the CTU Dataset. Moreover the categories do not overlap exactly; if an item has a collar it is categorized as polo in the CTU dataset even though it might look more like a jacket than a polo shirt to a human. Furthermore, the CTU dataset distinguishes between long and short sleeve items, whereas DeepFashion does not (\eg. \textit{tshirt} and \textit{tshirt-long} can both be in the \textit{Tee} category). We combine the categories as follows: bluse=(Blouse), hoody=(Hoodie, Sweater), pants=(Jeans, Jeggins, Joggers, Leggins), polo=(Tee, Button-Down), polo-long=(Button-Down, Henley, Jacket), skirt=(Skirt), tshirt=(Tee), tshirt-long=(Cardigan, Sweater, Tee). Since the DeepFashion dataset does not contain any towels, we ignore them in these experiments. 

For the second experiment, we split the images randomly into a \textit{train}, \textit{validate}, and \textit{test} set. (\ie 787, 240, 270 images).  

Both experiments are compared to the publicly available implementation of \citeauthor{Liu2018DFAnalysis} \cite{Liu2018DFAnalysis}. For a fair comparison we train both models with the same augmentation methods (\ie no augmentation, elastic warping (EW), rotation (R), and rotation  \& elastic warping (R \& EW)). For implementation details, please see the Appendix.

\begin{table*}[!ht]
\footnotesize

  \begin{adjustbox}{width=1\textwidth}
		\begin{tabular}{l|cccccccc|c}
      Methods (Trained on DF) & L.Collar & R.Collar & L.Sleeve & R.Sleeve & L.Waistline & R.Waistline & L.Hem & R.Hem & Avg. \\
      \hline
      \citeauthor{Liu2018DFAnalysis} \cite{Liu2018DFAnalysis} & 0.0819 & 0.1061 & \textbf{0.0910} & 0.0975 & 0.0185 & 0.0175 & 0.0437 & 0.0788 & 0.0669 \\
      
      Ours & \textbf{0.0557} & \textbf{0.0682} & 0.0947 & 0.1234 & \textbf{0.0177} & \textbf{0.0135} & 0.0497 & 0.0908 & 0.0642\\ 
      
      \citeauthor{Liu2018DFAnalysis} \cite{Liu2018DFAnalysis} EW & 0.0910 & 0.1059 & 0.0915 & \textbf{0.0470} & 0.0341 & 0.0196 & \textbf{0.0405} & 0.0690 & \textbf{0.0623} \\
      
      Ours EW & 0.0698 & 0.0923 & 0.1193 & 0.0843 & 0.0380 & 0.0315 & 0.0458 & \textbf{0.0525} & 0.0667\\ 
      
      \hline
      \citeauthor{Liu2018DFAnalysis} \cite{Liu2018DFAnalysis} R  & 0.0620 & 0.0930 & 0.0924 & 0.0663 & \textbf{0.0139} & 0.0171 & \textbf{0.0478} & 0.1035 & 0.0620 \\
      
      Ours R &  0.0621 & \textbf{0.0767} & 0.0949 & 0.0576 & 0.0527 & \textbf{0.0134} & 0.0926 & 0.0998 & 0.0687\\
      
      \citeauthor{Liu2018DFAnalysis} \cite{Liu2018DFAnalysis} R \& EW  & 0.0657 & 0.1135 & 0.0892 & \textbf{0.0523} & 0.0163 & 0.0206 & 0.0586 & \textbf{0.0662} & 0.0603 \\
      
      Ours R  \& EW& \textbf{0.0532} & 0.1129 & \textbf{0.0827} & 0.0535 & 0.0155 & 0.0202 & 0.0524 & 0.0817 & \textbf{0.0590}\\
      \hline

    \end{tabular}
    \end{adjustbox}
    \caption{Results on in-lab dataset for landmark localization on unknown items of clothing. The values represent the normalized error (NE). Best result marked in bold.}
  \label{table:inlab_LM}
\end{table*}

\begin{table*}[!ht]
\footnotesize
  \centering
		\begin{tabular}{l|cccccc|c}
      Methods & Hoodie & Jacket & Sweater & Tank & Tee & Jeans & Overall \\
      \hline
      \citeauthor{Liu2018DFAnalysis} \cite{Liu2018DFAnalysis} & 00.00 & \textbf{100.0} & 84.21 & \textbf{100.0} & 55.56 & \textbf{100.0} & 71.65 \\
      Ours & 05.88 & 88.89 & \textbf{100.0} & \textbf{100.0} & 62.96 & 96.30 & 76.07\\
      \citeauthor{Liu2018DFAnalysis} \cite{Liu2018DFAnalysis} R  & 00.00 & 11.11 & \textbf{100.0} & 77.78 & 62.96 & \textbf{100.0} & 66.67 \\
      Ours R & 00.00 & \textbf{100.0} & \textbf{100.0} & \textbf{100.0} & 62.96 & \textbf{100.0}  & 76.92 \\
      \citeauthor{Liu2018DFAnalysis} \cite{Liu2018DFAnalysis} R \& EW & 00.00 & 77.78 & \textbf{100.0} & \textbf{100.0} & \textbf{66.67} & \textbf{100.0} & 76.07\\
      Ours R \& EW & \textbf{47.06} & 88.89 & 94.74 & \textbf{100.0} & 51.85 & 96.30 & \textbf{78.63}  
      
    \end{tabular}
    \caption{Classification accuracy on in-lab dataset for unknown items of clothing. Best result marked in bold.}
  \label{table:inlab_cat}
  \vspace{-0.2cm}
\end{table*}

\subsubsection{Performance Evaluation}
The results of landmark prediction and category classification on the CTU dataset with pre-trained models are shown in Table \ref{table:CTU_LM_DF} (top) and \ref{table:CTU_cat_DF} respectively. The benefit of training with rotated images becomes apparent. This is not surprising since the pictures of garments in the CTU dataset are taken in any possible orientation, whereas in the DeepFashion dataset all items of clothing are upright. Adding elastic warping increases the performance further for the landmark prediction for all cases except the one where training was performed on DeepFashion with no rotation. Additional results, presented in the appendix, show that adjusting the parameters of elastic warping can improve the performance further in some cases.  Since it does achieve the best performance in the top-$3$ accuracy, we believe that a extended tuning of the elastic warping parameters could  therefore increase the landmark prediction as well as classification performance. The overall classification accuracy of $85\%$ shows that our model is able to generalize well even when trained on a dataset with significantly different configurations (\eg. items of clothing worn by persons) compared to $56\%$ reached by  \citeauthor{Liu2018DFAnalysis} \cite{Liu2018DFAnalysis}.


The results of the second experiment, trained and evaluated on CTU dataset, are shown in \cref{table:CTU_LM_DF} (bottom). Note that landmark predictions are significantly better when learned on the original dataset. 
The elastic warping seems to especially boost the performance in the case of no rotations. This is probably connected to the dataset composition and size as the EW augmented images boost the performance.

We omit the classification results since all the tested models achieve 100\% accuracy. 

Adding elastic warping as a data augmentation method can improve the performance in most of the evaluated cases. Our network outperforms the one proposed by \citeauthor{Liu2018DFAnalysis} \cite{Liu2018DFAnalysis} when trained with the same augmentation methods in both experiments. This indicates that state-of-the-art methods are likely to not generalize well to more challenging datasets.


\subsection{Experiments on In-Lab Dataset}
\label{ssec:experiment_robot}
In this experiment we analyze the inference performance of our network, solely trained on the DeepFashion dataset, on the images taken in a lab environment.  For implementation details, please see the Appendix.


The results for landmark prediction and category classification are shown in Table \ref{table:inlab_LM} and \ref{table:inlab_cat} respectively. Some landmark predictions are exemplified in Figure \ref{fig:inlab_dataset}. There is one item (\ie a hoody) that is in almost always misclassified, except when using elastic warping with our model. Furthermore, the long sleeve t-shirt (Figure \ref{fig:inlab_dataset} top row in the middle) is often classified as a sweater. With these two challenging items the best accuracy we achieve is $78.63\%$. Without these two items the accuracy increases to $93.33\%$. Due to the limited size of our dataset these two items have a significant impact. As the dataset is very limited in size, elastic warping can have a negative effect as well, as can be seen, for instance, in the drop in classification accuracy for the class Jacket. Further adjustments of the elastic warping parameters to achieve a higher similarity from the base images towards the task image might improve the results. Nevertheless, the combination of rotation and elastic warping leads to the best overall performance. The results of the landmark localization also show that our network is able to perform well even when an image contains parts of the robot.
This can be seen in the video provided in the supplementary material\footref{fn:code}, where a garment is being folded and the robotic arms occlude large parts of it. 

Elastic warping improves the performance for \citeauthor{Liu2018DFAnalysis} and leads to the best  performance of our network in both the landmark localization and the classification. This indicates that the augmentation helps the models to generalize between the datasets. Furthermore, the model adjustments, such as rotation invariant convolutions, improve the state of the art methods \cite{Liu2018DFAnalysis} and make an important step towards there usage in robotic clothing manipulation.

\section{Conclusion and Future Work}
\label{sec:conclusion}
In this work we use a large publicly available fashion image dataset with data augmentation to train a network for garment classification and landmark detection for robotic manipulation application. This is the first work where a deep learning model trained on RGB images is used for clothing category classification and landmark detection for robotic applications. We show that our model is able to generalize to robotic specific item configurations which differ significantly from the training dataset. We achieve this by utilizing rotation and our newly proposed \textit{elastic warping} augmentation method during training. 
After training, we use different datasets to evaluate the performance of our network.  We observe that other state-of-the-art methods, while producing excellent results on the training set, are not able to generalize as well to novel datasets. 


\section{Acknowledgements}
This work has been supported by the Swedish Foundation for Strategic Research, Swedish Research Council and Knut and Alice Wallenberg Foundation.

\section{Appendix}

In this appendix we give detailed account of the network structure and the experimental implementation details.

\subsection{Network structure}

We here describe the individual parts of the network in more detail.

\subsubsection{Rotation invariance}

In order to account for rotation, we replace the $2$D convolution in the layers \textit{conv\textsl{1}} to \textit{conv\textsl{4}} of the VGG16 network with \acp{A-ORConv} layers. These produce enriched feature maps with the orientation information explicitly encoded \cite{Wang2018IORN}. \acp{A-ORConv} are an improvement of the \acp{ORConv} initially proposed in \cite{Zhou2017ORN}. These convolution blocks use \acp{A-ARF} and \acp{ARF}, respectively. Both are a $5$D tensors of size $n_O \times n_I \times w_f \times h_f \times N$, where $n_O$ is the number of output channels, $n_I$ the number of input channels, $w_f$ and $h_f$ are the width and height of the filter and $N$ is the number of filter orientations. This means that in \acp{ARF} for each materialized filter, $N-1$ immaterialized rotated copies of the same filter are present. Therefore, during forward propagation one \ac{ARF} produces a feature map of $N$ channels with orientation information encoded. Depending on the orientation of the input image a different copy of the filter has the highest response. The improvement of \acp{A-ORConv} over \acp{ORConv} comes from reducing the risk of gradient explosion during training by updating the feature map with the mean value of the gradients from all its rotated copies instead of the sum of all gradients.

In our network, we use the \acp{A-ORConv} with four orientation channels (i.e. $N=4$). We use the same filter size and the same number of total channels when replacing the standard $2$D convolution in the \textit{conv\textsl{1}} to \textit{conv\textsl{4}} layers. This means that the effective number of parameters of the \acp{A-ORConv} is only a quarter of the normal convolution blocks. 

\begin{figure*}[ht]
  \centering
  \includegraphics[width=0.8\textwidth]{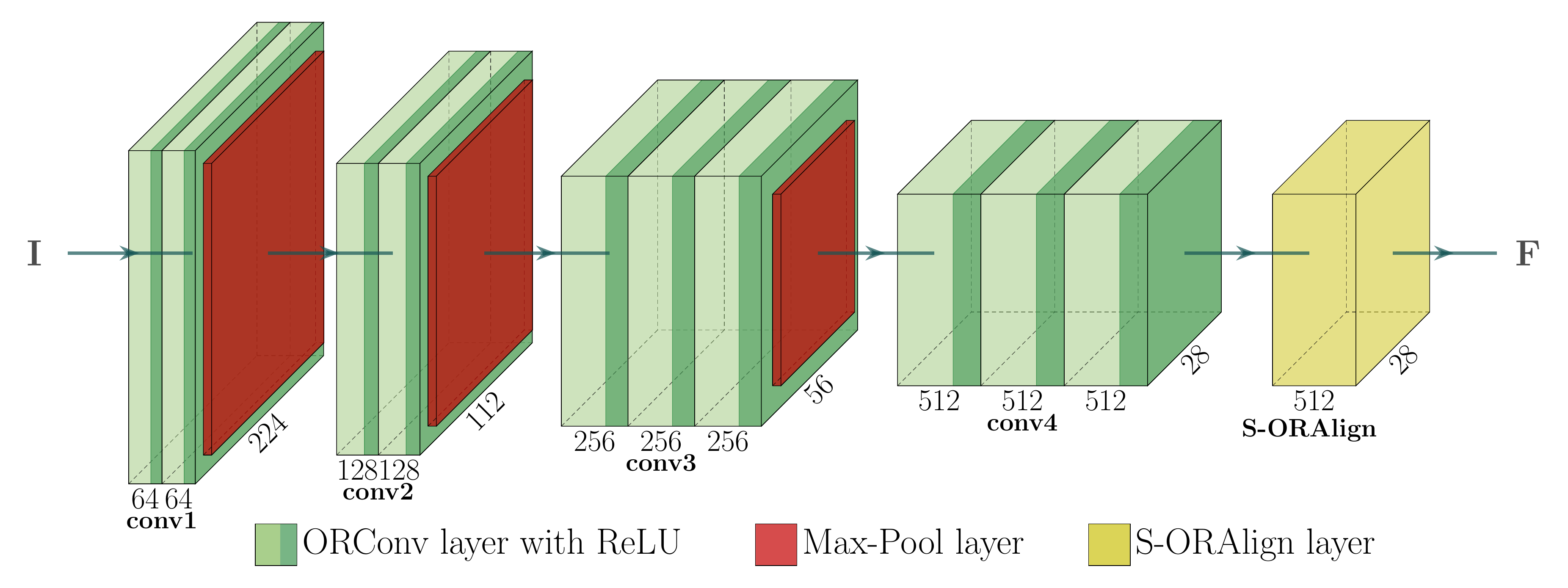}
  \caption{Structure of the landmark localization branch. Each cuboid represents a feature map of the given layer. The number below a cuboid denotes the number of channels in the feature map and the number on the side denotes the width and height of the feature map. (Best viewed in color)}
  \label{fig:orientation}
\end{figure*}

In order to get rotation invariant features S-ORAlign, proposed in \cite{Wang2018IORN}, is used to find the main response channel. The S-ORAlign is inspired by the Squeeze-and-Excitation (SE) block \cite{Hu2018Squeeze}, first a \textit{squeeze} operation is performed by global average pooling. Then the main orientation channel is found via a maximum function and finally all channels are spun such that the main response channel is in the first position.  The whole structure is depicted in Figure \ref{fig:orientation}.

\subsubsection{Landmark Localization Branch}
\label{ssec:landmark_localization_branch_a}
\begin{figure*}
\begin{subfigure}[b]{1\linewidth}
   \centering
  
 \includegraphics[width=0.7\linewidth]{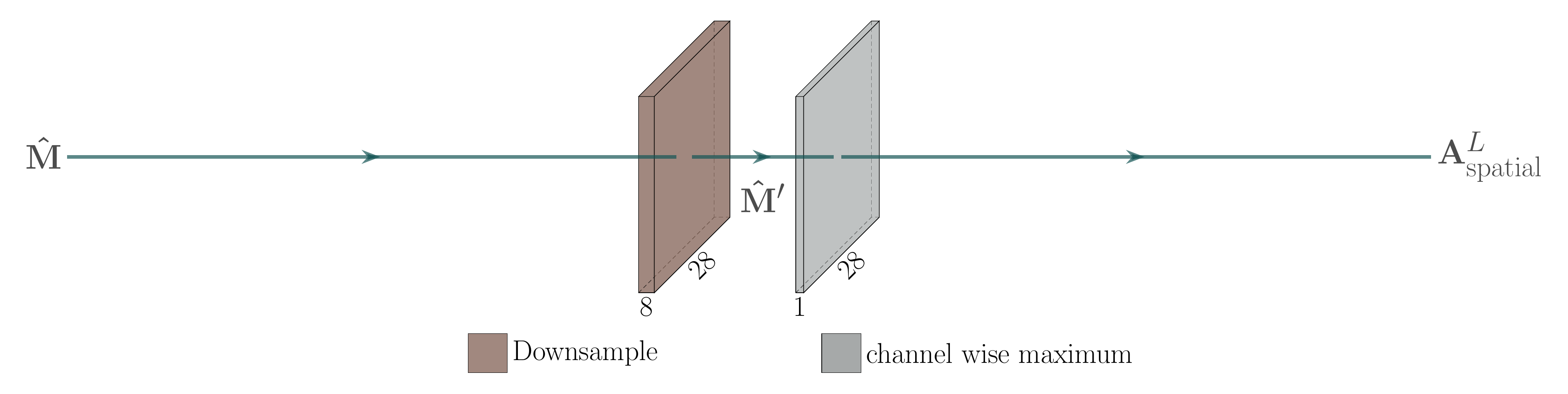}  
  \caption{Landmark aware spatial attention} \label{fig:attentiona}
 
   \includegraphics[width=0.7\linewidth]{figures/tex/category_attention.pdf}
  \caption{Category aware spatial attention} \label{fig:attentionb}
    \includegraphics[width=0.7\linewidth]{figures/tex/channel_attention.pdf}
   \caption{Channel attention} \label{fig:attentionc}
   \includegraphics[trim=100 0 0 0, width=0.7\linewidth]{figures/tex/attention_branch.pdf}
 \caption{Overall attention branch}\label{fig:attentiond}
\end{subfigure}
\caption{The different components of the attention branch. The number below a cuboid denotes the number of channels in the feature map and the number on the side denotes the width and height of the feature map. (Best viewed in color)}\label{fig:attention_branch}

\end{figure*}

The landmark localization branch is the same as proposed in \cite{Liu2018DFAnalysis}. The branch structure is depicted in Figure \ref{fig:attention_branch}. It uses transposed convolutions \cite{Dumoulin2016AGuide} to produce heatmaps for all landmarks. The transposed convolutions allow for an upsampling of the S-ORAlign features $\mat F \in \mathbb{R}^{28\times28\times512}$ back to the original input image size. Given the features $\mat F$ a $1\!\times\!1$ convolution is applied to reduce the number of channels in the feature map to $\mat F_L^{(1)} \in \mathbb{R}^{28\times28\times64}$. Then three blocks of two $3\!\times\!3$ convolutions followed by a $4\!\times\!4$ transposed convolution are utilized. The padding and stride of the transposed convolution are $1$ and $2$, respectively. Hence, such a block upsamples the feature maps by a factor of two, at the same time the number of channels is reduced by a factor of two. Finally a $1\!\times\!1$ convolution with a sigmoid activation is used to convert the $\mat F_L^{(4)}\in \mathbb{R}^{224\times224\times16}$ feature map into the predicted heatmaps $\hmat M \in [0,1]^{224\times224\times8}$.

The landmark localization branch can be trained separately from the classification. Let  $\mat M_k \in [0,1]^{224\times224}$ and $\hmat M_k \in [0,1]^{224\times224}$ denote the groundtruth heatmap and the predicted heatmap for the $k$th landmark, respectively. The landmark localization branch is trained using pixel-wise mean square differences,
\begin{align}
  \mathcal{L}_{\text{LM}} =  \sum_{i=1}^{n_B} \sum_{k=1}^{8} \sum_{x=1}^{224} \sum_{y=1}^{224} \| \mat M_k^i(x,y) - \hmat M_k^i(x,y) \|_2^2,
\end{align}
where $n_B$ is the total number of training samples. The groundtruth heatmap $\mat M_k^i$ is generated by adding a $2$D Gaussian filter at the corresponding location $\mat L^i_k$. Given a sample $i$ the predicted coordinates for the $k$th landmark $\hmat L^i_k$ corresponds to the maximal value in the predicted heatmap,

\begin{equation}
  \hmat L^i_k \in \underset{(x,y) \in \{1,\dots,224\} \times \{1,\dots,224\}}{\mathop{\text{argmax}}} \hmat M^i_k(x,y).
\end{equation}
If there is more than one maximum per landmark one of them is chosen at random.

 \begin{figure}[]
  \centering
  \includegraphics[width=0.45\textwidth]{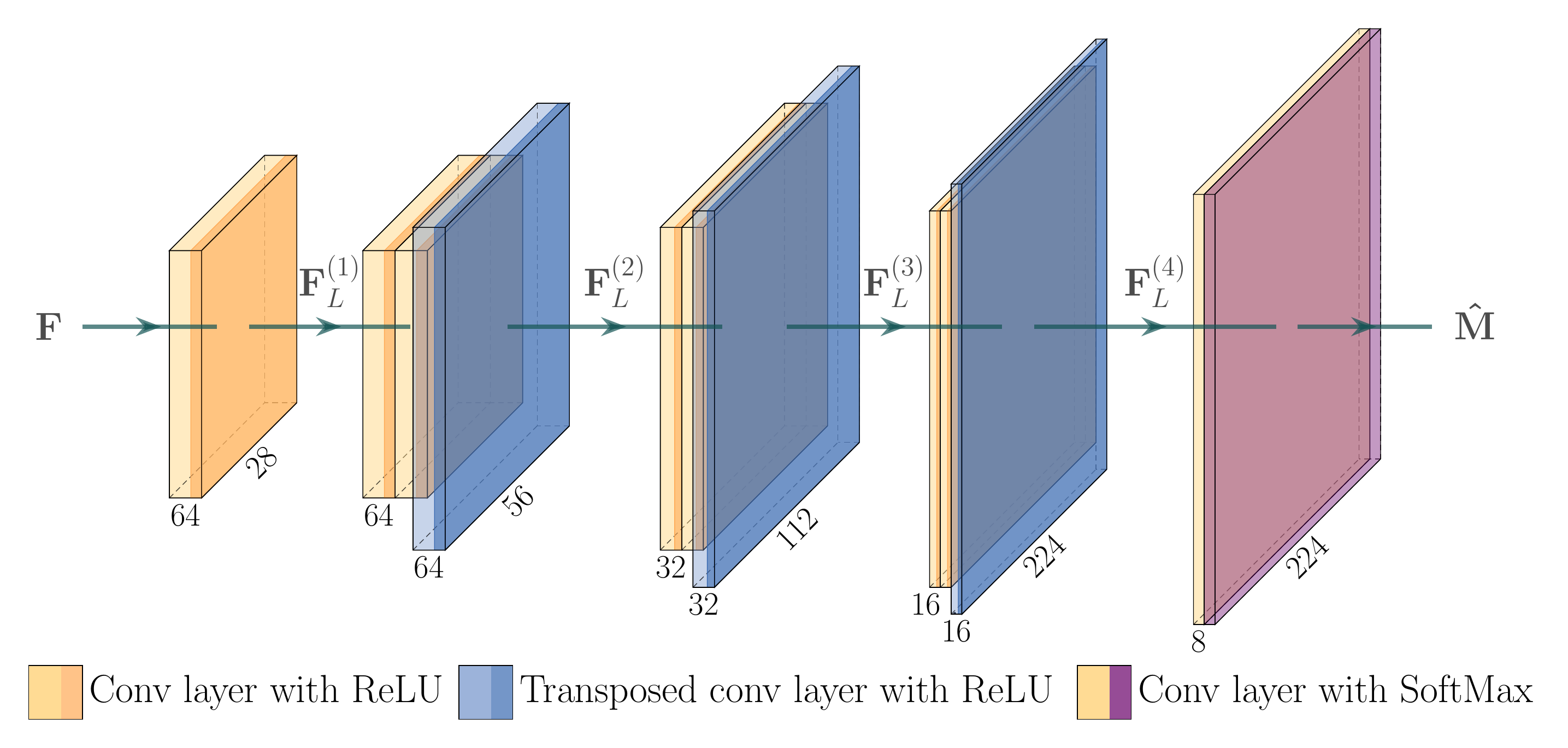}
  \caption{Structure of the landmark localization branch. Each cuboid represents a feature map of the given layer. The number below a cuboid denotes the number of channels in the feature map and the number on the side denotes the width and height of the feature map. (Best viewed in color)}
  \label{fig:LM_branch}
\end{figure}

 \subsubsection{Attention Branch}
\label{ssec:attention_branch_a}
The attention branch can be seen as a union of \textit{spatial} attention \cite{Jaderberg2015Spatial} and \textit{channel} attention \cite{Hu2018Squeeze}. The attention learns a saliency weight map $\mat A \in [-1, 1]^{28\times28\times512}$ of the same size as the S-ORAlign features $\mat F \in \mathbb{R}^{28\times28\times512}$. Inspired by the proposed attention modules in \cite{Wang2018Attentive} the spatial attention itself contains two types of attention, a landmark attention $\mat A^L_\text{spatial}$ and a category attention $\mat A^C_\text{spatial}$. 

We learn the spatial and channel attention in a factorized manner,
\begin{equation}
  \label{eq:attention_factorization_a}
  \mat A = (\mat A^L_\text{spatial}+\mat A^C_\text{spatial}) \times \mat A_\text{channel}.
\end{equation}

The attention branch is designed in a three branch unit; two branches for the spatial attention $\mat A_\text{spatial}^L, \mat A_\text{spatial}^C$ (Figure \ref{fig:attentiona}, Figure \ref{fig:attentionb}) and one for the channel attention $\mat A_\text{channel}$ (Figure \ref{fig:attentionc}). With the factorization (equation \ref{eq:attention_factorization_a}) combining them at the end, Figure \ref{fig:attentiond}.

\paragraph{Spatial Attention - Landmark}
Clothing landmarks represents functional regions of clothing and providing useful information about the item. The predicted heatmaps $\{\hmat M_k\}_{k=1}^8$ are used to get attention on the functional clothing regions. The weight map is created by downsampling the predicted heatmaps to the same size as the feature map in $\mat F$, followed by a max-pooling operation.

\begin{equation}
  \hmat M' = \left\{\operatorname{downsample}_{\,8}\:\hmat M_k\right\}_{k=1}^8
\end{equation}
\begin{align}
  \begin{gathered}
  \mat A^L_\text{spatial}(x,y) = \max_{k} \hmat M'_k(x,y) \\ 
  \quad \forall (x,y) \in \{1,\dots,28\} \times \{1,\dots,28\}
\end{gathered}
\end{align}
This attention is learned in a supervised manner since it is directly derived from the predicted heatmaps.

\paragraph{Spatial Attention - Category}
\label{p:category_attention}
Since the landmark attention only covers corner points of a clothing item, an additional spatial attention is used that focuses more on the clothing center. The category attention is modeled using an U-Net structure \cite{Ronneberger2015UNet}. 
Given the S-ORAlign features $\mat F \in \mathbb{R}^{28\times28\times512}$ a $1\!\times\!1$ convolution is applied to convert the features into $\mat F_A^{(1)} \in \mathbb{R}^{28\times28\times32}$. The U-Net consists of a contracting path that consists of two $4\times4$ convolutions with stride 2, which squeeze the features down to $\mat F_A^{(3)} \in \mathbb{R}^{7\times7\times128}$. The number of feature channels doubles at every contracting step. Then a $1\!\times\!1$ convolution and $4\times\!4$ transposed convolution are applied generating the features $\mat F_A^{(4)} \in \mathbb{R}^{14\times14\times128}$. Followed by the U-Net expanding path, which consists of two $4\times4$ transposed convolution. The input of the transposed convolution is a concatenation of the output from the previous transposed convolution and the corresponding feature map from the contraction path. The number of feature channels halves at every expanding step. At the end a $1\!\times\!1$ convolution is used to convert the channels to the same number as in the S-ORAlign features.

The downpooling to a low resolution of $7\!\times\!7$ gives the spatial attention a large receptive field in the feature map of $\mat F$. The up-sampling is then used to have a weight map of the same size as $\mat F$.
The model learns by itself which regions of an image are important. This is in contrast to our landmark attention, where the groundtruth heatmaps $\mat M$, which resemble the landmark attention, are provided during training.

\paragraph{Channel Attention}
The channel attention is implemented via a Squeeze-and-Excitation block \cite{Hu2018Squeeze}. First a \textit{squeeze} operation creates $\mat S \in \mathbb{R}^{512}$, an embedding of the global distribution of the channel-wise feature responses in $\mat F$. This channel descriptor is created using average pooling
\begin{equation}
  S(c) = \frac{1}{28\times28}\sum_{u=1}^{28} \sum_{v=1}^{28} \mat F(u,v,c) \quad \forall c \in \{1,\dots,512\}
\end{equation}
where $\mat F(\cdot,\cdot,c)$ is the feature map of the $c$th channel.

Then an \textit{excitation} operation is performed on the channel wise aggregated feature map to create the channel attention. 
\begin{equation}
  \mat A_{\text{channel}} = \sigma \left( \mat W_2 \operatorname{ReLU}(\mat W_1 \mat S) \right),
\end{equation}
where $\mat W_1 \in \mathbb{R}^{\frac{512}{r} \times 512}$, $\mat W_2 \in \mathbb{R}^{ 512\times\frac{512}{r}}$, and $\sigma$ represents the \textit{sigmoid} activation function. Following the proposal in \cite{Hu2018Squeeze} a bottleneck is created using two fully-connected layers, with a reduction rate $r$. We choose $r=16$ in all our experiments.

\paragraph{Factorization}
\label{p:factorization}
The factorization is then performed by multiplying the channel-wise feature responses in the spatial attention with the corresponding channel weights,
\begin{equation}
  \begin{gathered}
    \tmat A(x,y,c) = \left(\mat A^L_\text{spatial}(x,y,c)+\mat A^C_\text{spatial}(x,y,c)\right) \mat A_\text{channel}(c) \\
  \forall (x,y) \in \{1,\dots,28\}\times\{1,\dots,28\} \quad \forall c \in \{1,\dots,512\}.
\end{gathered}
\end{equation}
To refine the attention, an additional $1\!\times\!1$ convolution layer is added afterwards. This is motivated by the fact that the spatial and channel attention are not mutually exclusive but with co-occurring complementary relationship \cite{Li2018Harmonious}.

Afterwards, a \textit{tanh} function is used to shrink the attention values into a range of $\mat A \in [-1,1]^{28\times28\times512}$.

\subsubsection{Rest of the network}

Given $\mat A \in [-1,1]^{28\times28\times512}$ we weight the S-ORAlign features $\mat F \in \mathbb{R}^{28\times28\times512}$, 
\begin{equation}
  \mat U = (\mat 1+\mat A) \circ \mat F,
\end{equation}
where $\circ$ denotes the Hadamard product and $\mat 1$ is a tensor of ones with size $28\times28\times512$. Hence, features where $\mat A(\cdot, \cdot, \cdot) \in [-1, 0)$ are reduced and features where $\mat A(\cdot, \cdot, \cdot) \in (0, 1]$ are increased. Our attention incorporates semantic information and global information into the network helping to focus on important regions in the images. The features $\mat U$ are then fed in to the \textit{conv\textsl{5-1}} layer. The rest of the network follows the VGG-16 structure but with a reduced number of weights in the two fully connected layers (i.e. $1024$ and $1000$ instead of $4096$ and $1000$).

\begin{table}[ht]
  \begin{tabular}{ll}
    Method & best epoch  \\
    \hline
    Ours&   3\\ 
    Ours rot.& 47\\
    Ours rot. \& el. warp. & 50 
  \end{tabular}
  \caption{Number of epochs until early stopping (\ie best result on validation set).}
	\label{table:DeepFashion_epochs}
\end{table}

 \begin{table*}[]
   \centering
   \begin{adjustbox}{width=1\textwidth}
   \begin{tabular}{l|cccccccc|c}
     Methods & L.Collar & R.Collar & L.Sleeve & R.Sleeve & L.Waistline & R.Waistline & L.Hem & R.Hem & Avg. \\
     \hline 
     FashionNet \cite{Liu2016DeepFashion} & 0.0854 & 0.0902 & 0.0973 & 0.0935 & 0.0854 & 0.0845 & 0.0812 & 0.0823 & 0.0872 \\
     DFA \cite{Liu2016Landmarks} & 0.0628 & 0.0637 & 0.0658 & 0.0621 & 0.0726 & 0.0702 & 0.0658 & 0.0663 & 0.0660 \\
     DLAN \cite{Yan2017Unconstrained} & 0.0570 & 0.0611 & 0.0672 & 0.0647 & 0.0703 & 0.0694 & 0.0624 & 0.0627 & 0.0643 \\
    \citeauthor{Wang2018Attentive} \cite{Wang2018Attentive}  & 0.0415 & 0.0404 & 0.0496 & \textbf{0.0449} & 0.0502 & 0.0523 & \textbf{0.0537} & \textbf{0.0511} & 0.0484 \\
     \citeauthor{Liu2018DFAnalysis} \cite{Liu2018DFAnalysis} & \textbf{0.0332} & \textbf{0.0346} & 0.0487 & 0.0519 & \textbf{0.0422} & \textbf{0.0429} & 0.0620 & 0.0639 & 0.0474 \\
     \hline
     Ours  & 0.0343 & {0.0348} & 0.0488 & {0.0509} & {0.0436} & {0.0445} & 0.0582 & {0.0608}  & {0.0470}\\
     Ours R & 0.0351 & 0.0354 & \textbf{0.0480} & 0.0491 & 0.0440 & 0.0448 & 0.0564 & 0.0589 & \textbf{0.0466}\\
     Ours R \&  EW & 0.0368 & 0.0383 & 0.0506 & 0.0517 & 0.0499 & 0.0524 & 0.0578 & 0.0610 & 0.0498 \\ 
   \end{tabular}
   \end{adjustbox}
   \caption{Results on DeepFashion dataset for landmark localization. The values represent the normalized error (NE). Best results are marked in bold}
 	\label{table:DeepFashion_LM}
 \end{table*}


\subsection{Experiments}

In this section we describe the implementation details for each of the experiments and present results for the DeepFasion dataset. 

\subsubsection{DeepFashion Experiments}
\label{ssec:experiment_deepfashion_a}
\paragraph{Implementation Details}
We build our network using the publicly available implementation\footnote{\url{https://github.com/fdjingyuan/Deep-Fashion-Analysis-ECCV2018}} of \citeauthor{Liu2018DFAnalysis}\cite{Liu2016DeepFashion} as a starting point. The cropped images are resized to the input size of the VGG-16 network (\ie $224\times224$). The \acp{A-ORConv} and normal convolution layers are pretrained on ImageNet \cite{ImageNet2009}. The fully connected layers of the VGG-16 network are replaced with two separate fully connected layer branches, one for the category classification and the other for the attribute prediction. We use cross entropy loss for the category classification. Due to the inbalance between positive and negative samples asymmetric weighted cross entropy loss is used for the attribute prediction. The batch size is $32$ and $64$ during training and validation, respectively. The model is trained using the Adam optimizer \cite{Kingma2015Adam} with an initial learning rate of $0.0002$, which is multiplied by a factor of $0.8$ every fifth epoch. The landmark detection branch is initially trained separately for $20$ epochs. The landmark prediction is then locked and the learning rate is reset. Without locking the landmark prediction accuracy would decrease significantly during the classification training. The category classification and attribute prediction are trained for up to 50 epochs. We perform early stopping on the validation set. Meaning we track the best result on the validation set and stop the training if the result does not improve over $5$ consecutive epochs. The model state that achieved the best result is then used in the evaluation on the \textit{test} set. We do not perform specific parameter tuning depending on the dataset and/or the augmentation method. Furthermore, Table \ref{table:DeepFashion_epochs} shows the actually best epoch tracked with our early stopping.

\paragraph{Experimental results on DeepFashion}

We compare our landmark prediction results to the following five models \cite{Liu2016DeepFashion,Liu2016Landmarks, Yan2017Unconstrained, Wang2018Attentive, Liu2018DFAnalysis} and the clothing category classification to these models \cite{Chen2012Describing, Huang2015CrossDomain, Liu2016DeepFashion, Lu2017Fully, Corbiere2017Leveraging, Wang2018Attentive, Liu2018DFAnalysis}. We copy the results in Table \ref{table:DeepFashion_LM} and \ref{table:DeepFashion_Cat} as they were presented in \cite{Liu2018DFAnalysis} and add our own results. We also show the results when using our proposed data augmentation methods during training.

One can see that we outperform all other system in the landmark localization task when no augmentation or rotation is used. This indicates that our rotation invariance network structure is generally beneficial. That is specially noticeable for \textit{left/right sleeves}. These are the parts of clothing that generally have the most variation between images. On the other hand the category classification is not as good compared to the other systems. We assume that increasing the number of channels in the \ac{A-ORConv} layers could increase the accuracy. This is because the actual number of feature channels is only a fourth due to the rotated copies. As we show in the experiments in the main paper, our pretrained network outperforms  \citeauthor{Liu2018DFAnalysis} \cite{Liu2018DFAnalysis} when tested on other datasets. This suggests that the state-of-the-art models are not able to generalize from the training dataset.  

One can also see that our introduced \textit{elastic warping} performs worse on this dataset. When trained on augmented data, the network spreads its computational power over more possible clothing configurations which might decrease performance on a certain configuration (the untransformed testing data).

\subsubsection{Elastic Warping parameters Experiments}
We run additional experiments for Lanmark detection trained on DeepFashion and CTU dataset evaluated on the CTU dataset. Table \ref{table:CTU_LM_DF_1} shows the result for $\alpha=150$ and $\sigma = 10$, table \ref{table:CTU_LM_DF_2} shows the result for $\alpha=100$ and $\sigma = 10$, and \ref{table:CTU_LM_DF_3} shows the result for $\alpha=200$ and $\sigma = 10$. We can clearly see that the EW helps to boost the performance for the R \& EW augmentation when trained on the DeepFashion net and for only EW and R \& EW for the CTU case. The best performance for when trained on the DeepFashion or CTU is achived with $\alpha=100$ and $\sigma = 10$.


\begin{table*}[!ht]
      \begin{adjustbox}{width=1\textwidth}
		\begin{tabular}{l|cccccccc|c}
      Methods (Trained on DF) & L.Collar & R.Collar & L.Sleeve & R.Sleeve & L.Waistline & R.Waistline & L.Hem & R.Hem & Avg. \\
      \hline 
      \citeauthor{Liu2018DFAnalysis} \cite{Liu2018DFAnalysis}  & 0.5056 & \textbf{0.4810} & 0.3288 & 0.2623 & 0.4908 & 0.4665 & 0.4047 & 0.4774 & 0.4272 \\
      Ours & \textbf{0.4972} & 0.4835 & \textbf{0.2846} & \textbf{0.2055} & 0.4870 & 0.4677 & 0.4069 & 0.4727 & \textbf{0.4131} \\ 
       
      \citeauthor{Liu2018DFAnalysis} \cite{Liu2018DFAnalysis} EW  & 0.5123 &	0.5039 &	0.3440 &	0.2644	 & \textbf{0.4749} &	0.5010 &	\textbf{0.4018} &	\textbf{0.4660} &	0.4335 \\
      
      Ours EW & 0.5048 &	0.4982 &	0.3116 &	0.2893 &	0.4796  &	\textbf{0.4386} &	0.4190 &	0.4708 &	0.4265 \\ 
      
     \hline
     \citeauthor{Liu2018DFAnalysis} \cite{Liu2018DFAnalysis} R &  \textbf{0.0947} & 0.1004 & 0.0814 & \textbf{0.0670} & 0.1215 & \textbf{0.1018} & 0.2196 & 0.2177 & 0.1255\\
      Ours R & 0.1056 & 0.1075 & 0.0763 & 0.0708 & 0.1133 & 0.1206 & 0.1756 & \textbf{0.1526} & 0.1153 \\
      
      \citeauthor{Liu2018DFAnalysis} \cite{Liu2018DFAnalysis} R \& EW & 0.1077 &	0.1017 &	0.0873 &	0.0743 &	0.1215 &	0.1298 &	0.2187 &	0.2225 &	0.1329 \\
      
      Ours R \& EW & 0.1075 &	\textbf{0.0970} &	\textbf{0.0718} &	0.0715 &	\textbf{0.0976} &	0.1083 &	\textbf{0.1505} &	0.1569 &	\textbf{0.1076} \\  
       
      \hline

      Methods (Trained on CTU)  & L.Collar & R.Collar & L.Sleeve & R.Sleeve & L.Waistline & R.Waistline & L.Hem & R.Hem & Avg. \\
      \hline 
      \citeauthor{Liu2018DFAnalysis} \cite{Liu2018DFAnalysis} & 0.0560 & 0.0484 & 0.0473 & 0.0572 & 0.0473 & 0.0560 & 0.1010 & 0.0929 & 0.0632 \\
      Ours & 0.0500 & 0.0801 & 0.0790 & 0.0745 & 0.0590 & 0.0713 & 0.0749 & 0.0853 & 0.0719 \\
      
       \citeauthor{Liu2018DFAnalysis} \cite{Liu2018DFAnalysis} EW  & 0.0395	& 0.0388 &	0.0448 &	0.0750 &	0.0452 &	0.0467 &	0.1064 &	0.0848 &	0.0602 \\
       
      Ours EW & \textbf{0.0263} &	\textbf{0.0336} &	\textbf{0.0273} &	\textbf{0.0361} &	\textbf{0.0431} &	\textbf{0.0407} &	\textbf{0.0483} &	\textbf{0.0512} &	\textbf{0.0383} \\ 
      
     \hline
     
      \citeauthor{Liu2018DFAnalysis} \cite{Liu2018DFAnalysis} R  & 0.0299 & 0.0314 & 0.0289 & 0.0335 & 0.0560 & 0.0402 & 0.0539 & 0.0460 & 0.0400\\
      Ours R & \textbf{0.0181} & \textbf{0.0194} & 0.0253 & \textbf{0.0192} & 0.0374 & 0.0382 & 0.0314 & 0.0383 & 0.0284 \\ 
      
     \citeauthor{Liu2018DFAnalysis} \cite{Liu2018DFAnalysis} R \& EW  & 0.0319 &	0.0314 &	0.0332 &	0.0443 &	0.0559 &	0.0494 &	0.0442 &	0.0620 &	0.0440 \\
      
      Ours R \& EW & 0.0282 &	0.0251 &	\textbf{0.0230} &	0.0291 &	\textbf{0.0179} &	\textbf{0.0256} &	\textbf{0.0293} &	\textbf{0.0285} &	\textbf{0.0258} \\  
       
      \hline
      
    \end{tabular}
    \end{adjustbox}
  \caption{Results on CTU dataset for landmark localization with different augmentation methods, when trained on the DeepFashion (DF) dataset (top) and in the CTU dataset (bottom). The values represent the normalized error (NE). Best results are marked in bold EW parameters: $\alpha = 150$ , $\sigma = 10$.}
	\label{table:CTU_LM_DF_1} \vspace{-0.2cm}
\end{table*}

\begin{table*}[!ht]
      \begin{adjustbox}{width=1\textwidth}
		\begin{tabular}{l|cccccccc|c}
      Methods (Trained on DF) & L.Collar & R.Collar & L.Sleeve & R.Sleeve & L.Waistline & R.Waistline & L.Hem & R.Hem & Avg. \\
      \hline 
      \citeauthor{Liu2018DFAnalysis} \cite{Liu2018DFAnalysis}  & 0.5056 & \textbf{0.4810} & 0.3288 & 0.2623 & 0.4908 & 0.4665 & 0.4047 & 0.4774 & 0.4272 \\
      Ours & \textbf{0.4972} & 0.4835 & \textbf{0.2846} & \textbf{0.2055} & 0.4870 & 0.4677 & 0.4069 & 0.4727 & \textbf{0.4131} \\ 
       
      \citeauthor{Liu2018DFAnalysis} \cite{Liu2018DFAnalysis} EW  & 0.5014 &	0.5027 &	0.3287 &	0.2938 &	0.4997 &	0.4837 &	\textbf{0.3970} &	0.4684 &	0.4344 \\
      
      Ours EW & 0.5003 &	0.4829 &	0.2915 &	0.2334 &	\textbf{0.4689} &	\textbf{0.4629} &	0.4146 &	\textbf{0.4638} &	0.4148 \\ 
      
     \hline
     \citeauthor{Liu2018DFAnalysis} \cite{Liu2018DFAnalysis} R &  \textbf{0.0947} & 0.1004 & 0.0814 & 0.0670 & 0.1215 & 0.1018 & 0.2196 & 0.2177 & 0.1255\\
      Ours R & 0.1056 & 0.1075 & 0.0763 & 0.0708 & 0.1133 & 0.1206 & 0.1756 & \textbf{0.1526} & 0.1153 \\
      
      \citeauthor{Liu2018DFAnalysis} \cite{Liu2018DFAnalysis} R \& EW & 0.0961 &	0.0986 &	0.0830 &	0.0672 &	0.1082 &	0.1011 &	0.2161 &	0.2054 &	0.1220 \\
      
      Ours R \& EW & 0.0981 &	\textbf{0.0904} &	\textbf{0.0689} &	\textbf{0.0618} &	\textbf{0.0838} &	\textbf{0.0963} &	\textbf{0.1530} &	0.1643 &	\textbf{0.1021} \\  
       
      \hline

      Methods (Trained on CTU)  & L.Collar & R.Collar & L.Sleeve & R.Sleeve & L.Waistline & R.Waistline & L.Hem & R.Hem & Avg. \\
      \hline 
      \citeauthor{Liu2018DFAnalysis} \cite{Liu2018DFAnalysis} & 0.0560 & 0.0484 & 0.0473 & 0.0572 & 0.0473 & 0.0560 & 0.1010 & 0.0929 & 0.0632 \\
      Ours & 0.0500 & 0.0801 & 0.0790 & 0.0745 & 0.0590 & 0.0713 & 0.0749 & 0.0853 & 0.0719 \\
      
       \citeauthor{Liu2018DFAnalysis} \cite{Liu2018DFAnalysis} EW  & 0.0395	& 0.0388 &	0.0448 &	0.0750 &	0.0452 &	0.0467 &	0.1064 &	0.0848 &	0.0602 \\
       
      Ours EW & \textbf{0.0261} &	\textbf{0.0252} &	\textbf{0.0264} &	\textbf{0.0268} &	\textbf{0.0330} &	\textbf{0.0444} &	\textbf{0.0536} &	\textbf{0.0480} &	\textbf{0.0354} \\ 
      
     \hline
     
      \citeauthor{Liu2018DFAnalysis} \cite{Liu2018DFAnalysis} R  & 0.0299 & 0.0314 & 0.0289 & 0.0335 & 0.0560 & 0.0402 & 0.0539 & 0.0460 & 0.0400\\
      Ours R & \textbf{0.0181} & \textbf{0.0194} & \textbf{0.0253} & \textbf{0.0192} & 0.0374 & 0.0382 & 0.0314 & 0.0383 & 0.0284 \\ 
      
     \citeauthor{Liu2018DFAnalysis} \cite{Liu2018DFAnalysis} R \& EW  & 0.0214 &	0.0246 &	0.0300 &	0.0285 &	0.0412 &	0.0376 &	0.0439 &	0.0485 &	0.0345 \\
      
      Ours R \& EW & 0.0216 &	0.0186 &	0.0275 &	0.0237 &	\textbf{0.0252} &	\textbf{0.0314} &	\textbf{0.0239} &	\textbf{0.0275} &	\textbf{0.0249} \\  
       
      \hline
      
    \end{tabular}
    \end{adjustbox}
  \caption{Results on CTU dataset for landmark localization with different augmentation methods, when trained on the DeepFashion (DF) dataset (top) and in the CTU dataset (bottom). The values represent the normalized error (NE). Best results are marked in bold EW parameters: $\alpha = 100$ , $\sigma = 10$.}
	\label{table:CTU_LM_DF_2} \vspace{-0.2cm}
\end{table*}

\begin{table*}[!ht]
      \begin{adjustbox}{width=1\textwidth}
		\begin{tabular}{l|cccccccc|c}
      Methods (Trained on DF) & L.Collar & R.Collar & L.Sleeve & R.Sleeve & L.Waistline & R.Waistline & L.Hem & R.Hem & Avg. \\
      \hline 
      \citeauthor{Liu2018DFAnalysis} \cite{Liu2018DFAnalysis}  & 0.5056 & \textbf{0.4810} & 0.3288 & 0.2623 & 0.4908 & 0.4665 & 0.4047 & 0.4774 & 0.4272 \\
      Ours & \textbf{0.4972} & 0.4835 & 0.2846 & \textbf{0.2055} & \textbf{0.4870} & 0.4677 & 0.4069 & 0.4727 & \textbf{0.4131} \\ 
       
      \citeauthor{Liu2018DFAnalysis} \cite{Liu2018DFAnalysis} EW  & 0.5098 &	0.4999 &	0.3227 &	0.2810 &	0.5012 &	0.4949 &	\textbf{0.3944} &	\textbf{0.4563} &	0.4325 \\
      
      Ours EW & 0.5174 &	0.4926 &	\textbf{0.2813} &	0.2372 &	0.4918 &	\textbf{0.4393} &	0.4366 &	0.4766	& 0.4216 \\ 
      
     \hline
     \citeauthor{Liu2018DFAnalysis} \cite{Liu2018DFAnalysis} R &  \textbf{0.0947} & 0.1004 & 0.0814 & 0.0670 & 0.1215 & \textbf{0.1018} & 0.2196 & 0.2177 & 0.1255\\
      Ours R & 0.1056 & 0.1075 & \textbf{0.0763} & 0.0708 & 0.1133 & 0.1206 & 0.1756 & \textbf{0.1526} & 0.1153 \\
      
      \citeauthor{Liu2018DFAnalysis} \cite{Liu2018DFAnalysis} R \& EW & 0.1008 &	\textbf{0.0901} &	0.0849 &	\textbf{0.0637} &	0.1205 &	0.1134 &	0.2144 &	0.2096 &	0.1247 \\
      
      Ours R \& EW & 0.0977 &	0.1058 &	0.0801 &	0.0643 &	\textbf{0.0920} &	0.1192 &	\textbf{0.1683} &	0.1747 &	\textbf{0.1128} \\  
       
      \hline

      Methods (Trained on CTU)  & L.Collar & R.Collar & L.Sleeve & R.Sleeve & L.Waistline & R.Waistline & L.Hem & R.Hem & Avg. \\
      \hline 
      \citeauthor{Liu2018DFAnalysis} \cite{Liu2018DFAnalysis} & 0.0560 & 0.0484 & 0.0473 & 0.0572 & 0.0473 & 0.0560 & 0.1010 & 0.0929 & 0.0632 \\
      Ours & 0.0500 & 0.0801 & 0.0790 & 0.0745 & 0.0590 & 0.0713 & 0.0749 & 0.0853 & 0.0719 \\
      
       \citeauthor{Liu2018DFAnalysis} \cite{Liu2018DFAnalysis} EW  & 0.0374 &	0.0404 &	0.0493 &	0.0627 &	0.0514 &	0.0390 &	0.0828 &	0.0818 &	0.0556 \\
       
      Ours EW & \textbf{0.0324} &	\textbf{0.0323} &	\textbf{0.0304} &	\textbf{0.0472} &	\textbf{0.0210} &	\textbf{0.0179} &	\textbf{0.0555} &	\textbf{0.0515} &	\textbf{0.0360} \\ 
      
     \hline
     
      \citeauthor{Liu2018DFAnalysis} \cite{Liu2018DFAnalysis} R  & 0.0299 & 0.0314 & 0.0289 & 0.0335 & 0.0560 & 0.0402 & 0.0539 & 0.0460 & 0.0400\\
      Ours R & \textbf{0.0181} & \textbf{0.0194} & 0.0253 & 0.0192 & 0.0374 & 0.0382 & 0.0314 & 0.0383 & 0.0284 \\ 
      
     \citeauthor{Liu2018DFAnalysis} \cite{Liu2018DFAnalysis} R \& EW  & 0.0241 &	0.0270 &	0.0268 &	0.0287 &	0.0411 &	0.0433 &	0.0498 &	0.0531 &	0.0367 \\
      
      Ours R \& EW & 0.0222 &	0.0239 &	\textbf{0.0157} &	\textbf{0.0180} &	\textbf{0.0356} &	\textbf{0.0302} &	\textbf{0.0212} &	\textbf{0.0348} &	\textbf{0.0252} \\  
       
      \hline
      
    \end{tabular}
    \end{adjustbox}
  \caption{Results on CTU dataset for landmark localization with different augmentation methods, when trained on the DeepFashion (DF) dataset (top) and in the CTU dataset (bottom). The values represent the normalized error (NE). Best results are marked in bold EW parameters: $\alpha = 200$ , $\sigma = 10$.}
	\label{table:CTU_LM_DF_3} \vspace{-0.2cm}
\end{table*}

 \begin{table}[]
   \begin{tabular}{lcc }
     \multirow{2}{*}{Methods} & \multicolumn{2}{c }{Category}   \\
                              & top-3 & top-5   \\
     \hline
     WTBI \cite{Chen2012Describing} & 43.73 & 66.26   \\
     DARN \cite{Huang2015CrossDomain} & 59.48 & 79.58   \\
     FashionNet \cite{Liu2016DeepFashion} & 82.58 & 90.17   \\
     \citeauthor{Lu2017Fully} \cite{Lu2017Fully} & 86.72 & 92.51 \\
     \citeauthor{Corbiere2017Leveraging} \cite{Corbiere2017Leveraging} & 86.30 & 93.80  \\
     \citeauthor{Wang2018Attentive} \cite{Wang2018Attentive}& 90.99 & 95.78   \\
     \citeauthor{Liu2018DFAnalysis} \cite{Liu2018DFAnalysis} & \textbf{91.16} & \textbf{96.12}   \\
     \hline
     Ours & 89.02 & 94.80 \\
     Ours R & 89.57 & 95.09    \\
     Ours R \& EW  & 89.63 & 95.10  \\ 
 \end{tabular}
   \caption{Results on DeepFashion dataset for clothing classification and attribute prediction values are in \%. Best results are marked in bold}
 	\label{table:DeepFashion_Cat}
 \end{table}

\begin{table}[]
  \begin{tabularx}{\linewidth}{lX}
      \textbf{CTU categories} & \textbf{DeepFashion categories} \\
      \hline
      bluse & Blouse\\
      hoody & Hoodie, Sweater \\
      pants & Jeans, Jeggins, Joggers, Leggins \\
      polo & Tee, Button-Down \\
      polo-long & Button-Down, Henley, Jacket \\
      skirt & Skirt \\ 
      tshirt & Tee \\ 
      tshirt-long & Cardigan, Sweater, Tee
    \end{tabularx}
  \caption{Mapping of clothing categories between the DeepFashion dataset and the CTU dataset.}
	\label{table:CTU_categories}
\end{table}

\subsection{Implementation details for the CTU experiments}

In the first experiment, we use the network trained as described in Section \ref{ssec:experiment_deepfashion_a} and perform solely inference with it. We do only consider the 13 categories in Table \ref{table:CTU_categories} as possible predictions and mask the others out.

The network setup for the second experiment is the same as for the DeepFashion dataset described in Section \ref{ssec:experiment_deepfashion_a} with the exception of the following changes. The last fully connected layer of the VGG-16 network is reduced from 1000 categories down to 9 categories. The number of epochs is increased to a maximum of $200$ since the dataset is much smaller and the learning rate decreases every $25$th epoch. The landmark prediction is initially trained for 50 epochs.

\subsubsection{Implementation details for the In-Lab Dataset}
In this experiment we use the network trained as described in Section \ref{ssec:experiment_deepfashion_a}. We perform solely inference with the network. For the evaluation we only consider the categories that are present and mask the rest out.

\bibliographystyle{IEEEtranN}

\bibliography{main}




\newpage

\end{document}